\documentclass[10pt,twocolumn,letterpaper]{article}

\usepackage{bm}
\usepackage{iccv}
\usepackage{times}
\usepackage{epsfig}
\usepackage{graphicx}
\usepackage{amsmath}
\usepackage{amssymb}
\usepackage{subfigure}
\usepackage{amsthm}
\usepackage{mathrsfs}
\usepackage{algorithmic}
\usepackage{verbatim}
\usepackage{booktabs}
\usepackage{colortbl}
\usepackage{caption}
\usepackage{authblk}
\usepackage{url}
\usepackage[linesnumbered,boxed]{algorithm2e}

\def\mycolor{\cellcolor[rgb]{0.8275,0.8275,0.8275}}

\usepackage[breaklinks=true,bookmarks=false]{hyperref}

\iccvfinalcopy 

\def\iccvPaperID{****} 

\begin{document}

\title{CenterNet: Keypoint Triplets for Object Detection}
\author{Kaiwen Duan$^{1}$\thanks{This work was done when the first author was interning at Huawei Noah's Ark Lab.}\qquad Song Bai$^{2}$\qquad Lingxi Xie$^{3}$\qquad Honggang Qi$^{1}$\qquad Qingming Huang$^{1}$\qquad Qi Tian$^{3}$\\
$^{1}$University of Chinese Academy of Sciences\qquad $^{2}$University of Oxford\qquad$^{3}$Huawei Noah's Ark Lab\\
{\tt\small kaiwen.duan@vipl.ict.ac.cn\qquad songbai.site@gmail.com\qquad 198808xc@gmail.com\qquad hgqi@ucas.ac.cn\qquad qmhuang@ucas.ac.cn \qquad tian.qi1@huawei.com}}
\maketitle

\begin{abstract}
In object detection, keypoint-based approaches often suffer a large number of incorrect object bounding boxes, arguably due to the lack of an additional look into the cropped regions. This paper presents an efficient solution which explores the visual patterns within each cropped region with minimal costs. We build our framework upon a representative one-stage keypoint-based detector named CornerNet. Our approach, named CenterNet, detects each object as a triplet, rather than a pair, of keypoints, which improves both precision and recall. Accordingly, we design two customized modules named cascade corner pooling and center pooling, which play the roles of enriching information collected by both top-left and bottom-right corners and providing more recognizable information at the central regions, respectively. On the MS-COCO dataset, CenterNet achieves an AP of \textbf{47.0\%}, which outperforms all existing one-stage detectors by at least \textbf{4.9\%}. Meanwhile, with a faster inference speed, CenterNet demonstrates quite comparable performance to the top-ranked two-stage detectors. Code is available at \url{https://github.com/Duankaiwen/CenterNet}.
\end{abstract}
\section{Introduction}
Object detection has been significantly improved and advanced with the help of deep learning, especially convolutional neural networks~\cite{girshick2014rich} (CNNs). In the current era, one of the most popular flowcharts is anchor-based~\cite{girshick2015fast,he2017mask,liu2016ssd,redmon2016you, ren2015faster}, which placed a set of rectangles with pre-defined sizes, and regressed them to the desired place with the help of ground-truth objects. These approaches often need a large number of anchors to ensure a sufficiently high IoU (intersection over union) rate with the ground-truth objects, and the size and aspect ratio of each anchor box need to be manually designed. In addition, anchors are usually not aligned with the ground-truth boxes, which is not conducive to the bounding box classification task.

\begin{figure}[tb]
\centering 
\subfigure{
\includegraphics[height=0.165\textwidth,width=0.165\textheight]{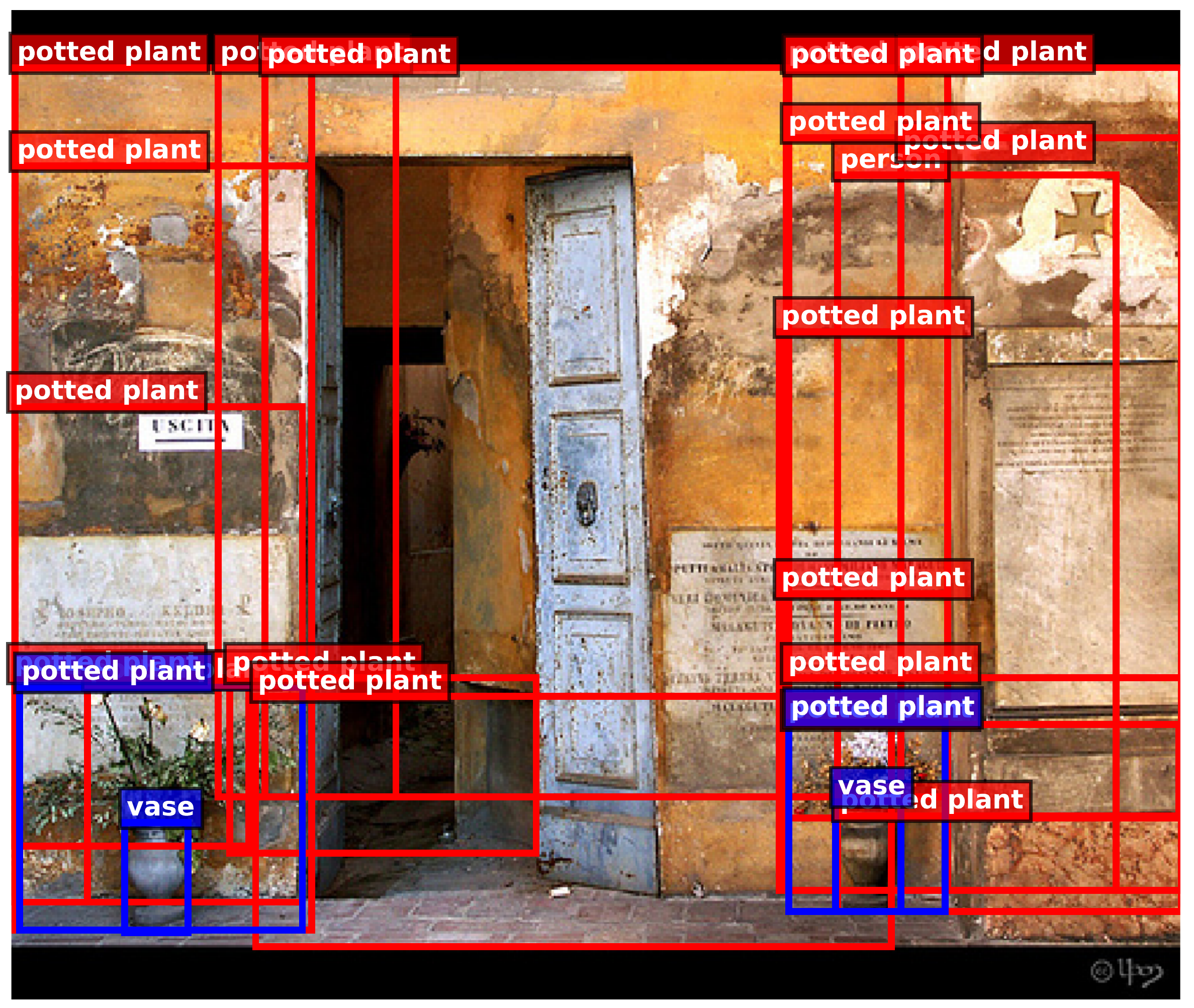}
\hspace{0.05in}
\includegraphics[height=0.165\textwidth,width=0.165\textheight]{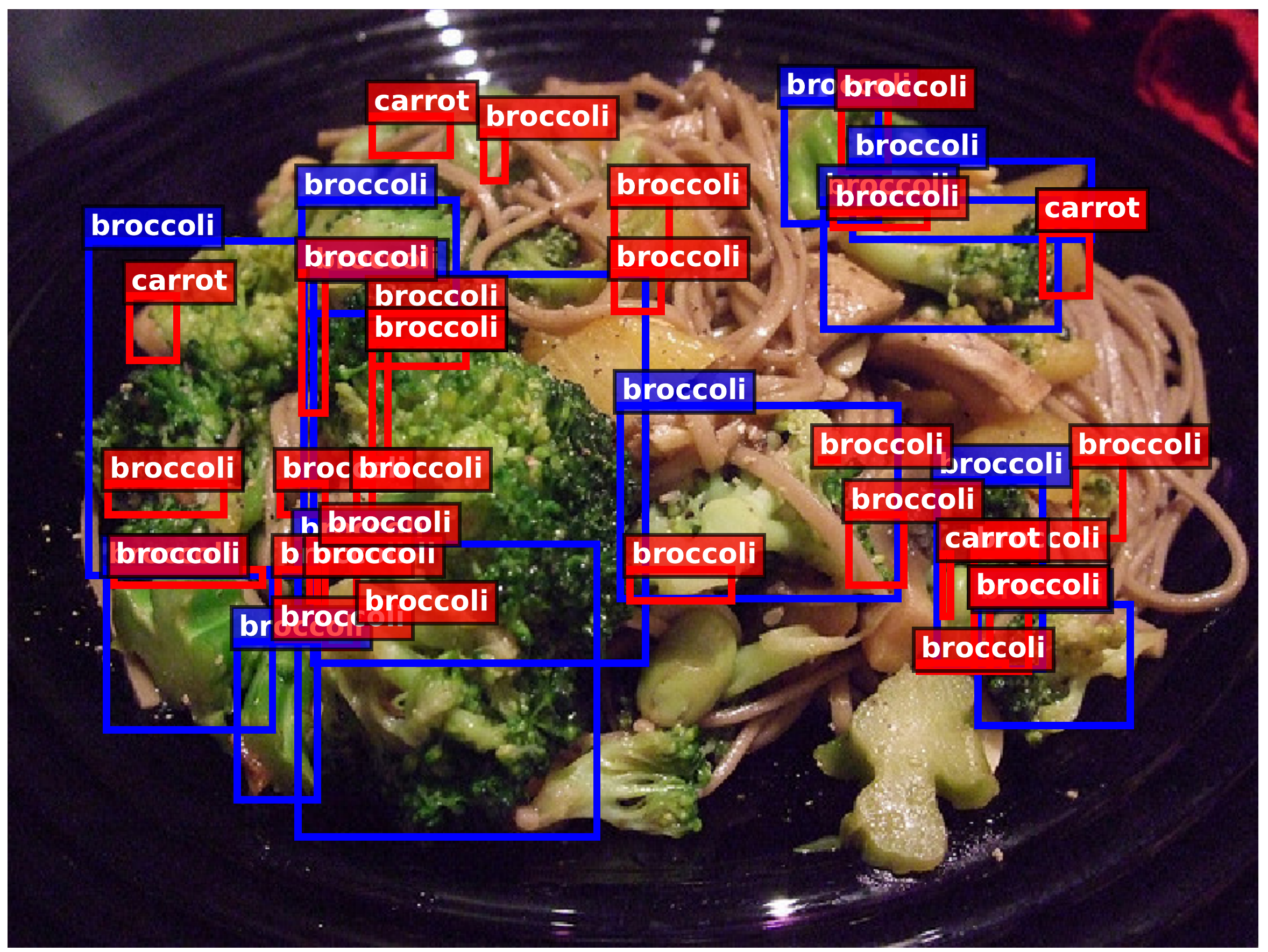}
\label{fig1a}
}
\subfigure{
\includegraphics[height=0.155\textwidth,width=0.345\textheight]{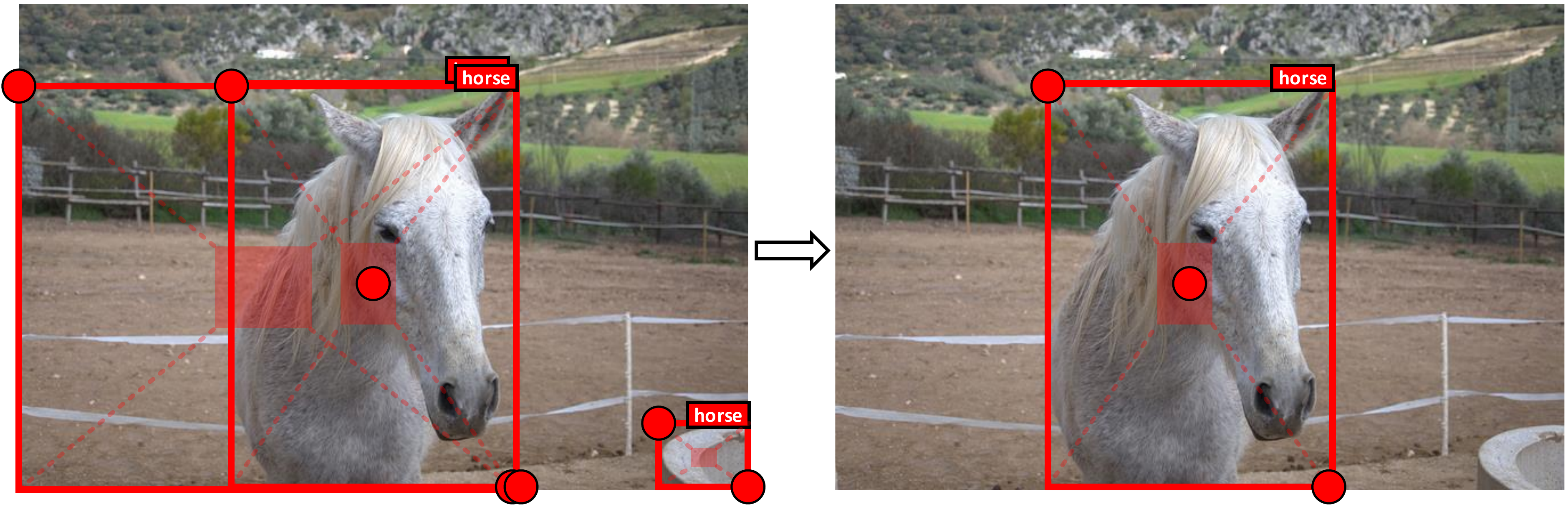}
\label{fig1b}
}
\vspace{-2ex}
\caption{In the first row, we visualize the top 100 bounding boxes (according to the MS-COCO dataset standard) of CornerNet. Ground-truth and predicted objects are marked in blue and red, respectively. In the second row, we show that correct predictions can be determined by checking the central parts.}
\label{fig1}
\end{figure}
To overcome the drawbacks of anchor-based approaches, a keypoint-based object detection pipeline named CornerNet~\cite{law2018cornernet} was proposed. It represented each object by a pair of corner keypoints, which bypassed the need of anchor boxes and achieved the state-of-the-art one-stage object detection accuracy. Nevertheless, the performance of CornerNet is still restricted by its relatively weak ability of referring to the global information of an object. That is to say, since each object is constructed by a pair of corners, the algorithm is sensitive to detect the boundary of objects, meanwhile not being aware of which pairs of keypoints should be grouped into objects. Consequently, as shown in Figure~\ref{fig1}, it often generates some incorrect bounding boxes, most of which could be easily filtered out with complementary information, {\em e.g.}, the aspect ratio.

To address this issue, we equip CornerNet with an ability of perceiving the visual patterns within each proposed region, so that it can identify the correctness of each bounding box by itself. In this paper, we present a low-cost yet effective solution named {\bf CenterNet}, which explores the central part of a proposal, {\em i.e.}, the region that is close to the geometric center, with one extra keypoint. Our intuition is that, if a predicted bounding box has a high IoU with the ground-truth box, then the probability that the center keypoint in its central region is predicted as the same class is high, and vice versa. Thus, during inference, after a proposal is generated as a pair of corner keypoints, we determine if the proposal is indeed an object by checking if there is a center keypoint of the same class falling within its central region. The idea, as shown in Figure~\ref{fig1}, is to use a triplet, instead of a pair, of keypoints to represent each object. 

Accordingly, for better detecting center keypoints and corners, we propose two strategies to enrich center and corner information, respectively.
The first strategy is named {\bf center pooling}, which is used in the branch for predicting center keypoints. Center pooling helps the center keypoints obtain more recognizable visual patterns within objects, which makes it easier to perceive the central part of a proposal. We achieve this by getting out the max summed response in both horizontal and vertical directions of the center keypoint on a feature map for predicting center keypoints. 
%
The second strategy is named {\bf cascade corner pooling}, which equips the original corner pooling module~\cite{law2018cornernet} with the ability of perceiving internal information. We achieve this by getting out the max summed response in both boundary and internal directions of objects on a feature map for predicting corners. Empirically, we verify that such a two-directional pooling method is more stable, {\em i.e.}, being more robust to feature-level noises, which contributes to the improvement of both precision and recall. 

We evaluate the proposed CenterNet on the MS-COCO dataset~\cite{lin2014microsoft}, one of the most popular benchmarks for large-scale object detection. CenterNet, with both center pooling and cascade corner pooling incorporated, reports an AP of $\mathbf{47.0\%}$ on the test-dev set, which outperforms all existing one-stage detectors by a large margin. With an average inference time of $270\mathrm{ms}$ using a 52-layer hourglass backbone~\cite{newell2016stacked} and $340\mathrm{ms}$ using a 104-layer hourglass backbone~\cite{newell2016stacked} per image, CenterNet is quite efficient yet closely matches the state-of-the-art performance of the other two-stage detectors.

The remainder of this paper is organized as follows. Section~\ref{RelatedWork} briefly reviews related work, and Section~\ref{Approach} details the proposed CenterNet. Experimental results are given in Section~\ref{Experiments}, followed by the conclusion in Section~\ref{Conclusions}.

\section{Related Work}
\label{RelatedWork}

Object detection involves locating and classifying the objects. In the deep learning era, powered by deep convolutional neural networks, object detection approaches can be roughly categorized into two main types of pipelines, namely, two-stage approaches and one-stage approaches.

\vspace{1ex}\noindent \textbf{Two-stage approaches}~divide the object detection task into two stages: extract RoIs, then classify and regress the RoIs.

R-CNN~\cite{girshick2014rich} uses a selective search method~\cite{uijlings2013selective} to locate RoIs in the input images and uses a DCN-based regionwise classifier to classify the RoIs independently. SPP-Net~\cite{he2015spatial} and Fast-RCNN~\cite{girshick2015fast} improve R-CNN by extracting the RoIs from the feature maps. Faster-RCNN~\cite{ren2015faster} is allowed to be trained end to end by introducing RPN (region proposal network). RPN can generate RoIs by regressing the anchor boxes. Later, the anchor boxes are widely used in the object detection task. Mask-RCNN~\cite{he2017mask} adds a mask prediction branch on the Faster-RCNN, which can detect objects and predict their masks at the same time. R-FCN~\cite{dai2016r} replaces the fully connected layers with the position-sensitive score maps for better detecting objects. Cascade R-CNN~\cite{cai2018cascade} addresses the problem of overfitting at training and quality mismatch at inference by training a sequence of detectors with increasing IoU thresholds. The keypoint-based object detection approaches~\cite{tychsen2017denet,Lu2018Grid} are proposed to avoid the disadvantages of using anchor boxes and bounding boxes regression. Other meaningful works are proposed for different problems in object detection, \eg,~\cite{zhu2017couplenet,lee2017me} focus on the architecture design,~\cite{bell2016inside,gidaris2015object,shrivastava2016contextual,zeng2016gated} focus on the contextual relationship,~\cite{li2019scale,cai2016unified} focus on the multi-scale unification.
\begin{figure*}[!tb]
  \centering 
  \includegraphics[width=0.98\textwidth]{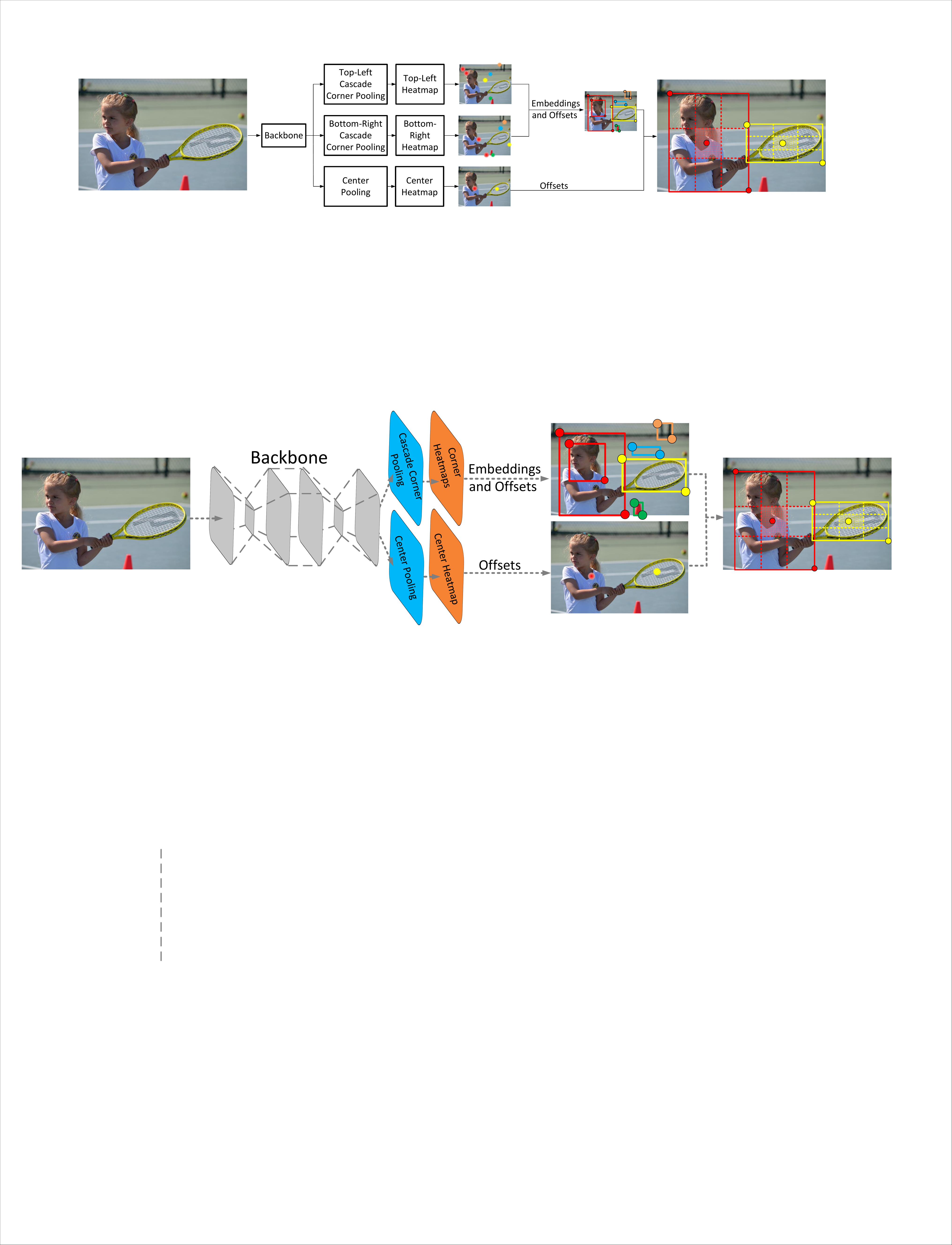}
  \vspace{-2ex}
  \caption{Architecture of CenterNet. A convolutional backbone network applies cascade corner pooling and center pooling to output two corner heatmaps and a center keypoint heatmap, respectively. Similar to CornerNet, a pair of detected corners and the similar embeddings are used to detect a potential bounding box. Then the detected center keypoints are used to determine the final bounding boxes.} 
  \label{structure} 
\end{figure*}

\vspace{1ex}\noindent \textbf{One-stage approaches}~remove the RoI extraction process and directly classify and regress the candidate anchor boxes.

YOLO~\cite{redmon2016you} uses fewer anchor boxes (divide the input image into an $\mathrm{S}\times\mathrm{S}$ grid) to do regression and classification. YOLOv2~\cite{redmon2017yolo9000} improves the performance by using more anchor boxes and a new bounding box regression method. SSD~\cite{liu2016ssd} places anchor boxes densely over an input image and use features from different convolutional layers to regress and classify the anchor boxes. DSSD~\cite{fu2017dssd} introduces a deconvolution module into SSD to combine low-level and high-level features. While R-SSD~\cite{jeong2017enhancement} uses pooling and deconvolution operations in different feature layers to combine low-level and high-level features. RON~\cite{kong2017ron} proposes a reverse connection and an objectness prior to extract multiscale features effectively. RefineDet~\cite{zhang2018single} refines the locations and sizes of the anchor boxes for two times, which inherits the merits of both one-stage and two-stage approaches. CornerNet~\cite{law2018cornernet} is another keypoint-based approach, which directly detects an object using a pair of corners. Although CornerNet achieves high performance, it still has more room to improve.

\section{Our Approach}
\label{Approach}

\subsection{Baseline and Motivation}\label{baseline}
This paper uses CornerNet~\cite{law2018cornernet} as the baseline. For detecting corners, CornerNet produces two heatmaps: a heatmap of top-left corners and a heatmap of bottom-right corners. The heatmaps represent the locations of keypoints of different categories and assigns a confidence score for each keypoint. Besides, it also predicts an embedding and a group of offsets for each corner. The embeddings are used to identify if two corners are from the same object. The offsets learn to remap the corners from the heatmaps to the input image. For generating object bounding boxes, top-$k$ left-top corners and bottom-right corners are selected from the heatmaps according to their scores, respectively. Then, the distance of the embedding vectors of a pair of corners is calculated to determine if the paired corners belong to the same object. An object bounding box is generated if the distance is less than a threshold. The bounding box is assigned a confidence score, which equals to the average scores of the corner pair. 
\begin{table}[tb]
\centering
\resizebox{.48\textwidth}{!}{
\begin{tabular}{|l|ccccccc|}
\hline
Method & FD & FD$_{5}$ & FD$_{25}$ & FD$_{50}$ & FD$_{\mathrm{S}}$ & FD$_{\mathrm{M}}$ & FD$_{\mathrm{L}}$\\
\hline
CornerNet & 37.8 & 32.7 & 36.8 & 43.8 & 60.3 & 33.2 & 25.1 \\
\hline
\end{tabular}}
\vspace{-2ex}
\caption{False discovery rates ($\%$) of CornerNet. The false discovery rate reflects the distribution of incorrect bounding boxes. The results suggest the incorrect bounding boxes account for a large proportion.}
\label{FDR1}
\end{table}

In Table~\ref{FDR1}, we provide a deeper analysis of CornerNet. We count the FD\footnote{$\mathrm{FD}=1-\mathrm{AP}$, where AP denotes the average precision at $\mathrm{IoU = [0.05 : 0.05 : 0.5]}$ on the MS-COCO dataset. Also, $\mathrm{FD}_{i} = 1 - \mathrm{AP}_{i}$, where $\mathrm{AP_i}$ denotes the average precision at $\mathrm{IoU} = i/100$, $\mathrm{FD_{scale}} = 1 - \mathrm{AP_{scale}}$, where $\mathrm{scale} = \left\{\mathrm{small, medium, large}\right\}$, denotes the scale of object.} (false discovery) rate of CornerNet on the MS-COCO validation dataset, defined as the proportion of the incorrect bounding boxes. The quantitative results demonstrate the incorrect bounding boxes account for a large proportion even at low IoU thresholds, \eg, CornerNet obtains $32.7\%$ FD rate at $\mathrm{IoU = 0.05}$. This means in average, $32.7$  out of every $100$ object bounding boxes have IoU lower than $0.05$ with the ground-truth. The small incorrect bounding boxes are even more, which achieves $60.3\%$ FD rate. One of the possible reasons lies in that CornerNet cannot look into the regions inside the bounding boxes. To make CornerNet~\cite{law2018cornernet} perceive the visual patterns in bounding boxes, one potential solution is to adapt CornerNet into a two-stage detector, which uses the RoI pooling~\cite{girshick2015fast} to look into the visual patterns in bounding boxes. However, it is known that such a paradigm is computationally expensive.

In this paper, we propose a highly efficient alternative called \textbf{CenterNet} to explore the visual patterns within each bounding box.
For detecting an object, our approach uses a triplet, rather than a pair, of keypoints. By doing so, our approach is still a one-stage detector, but partially inherits the functionality of RoI pooling. Our approach only pays attention to the center information, the cost of our approach is minimal. Meanwhile, we further introduce the visual patterns within objects into the keypoint detection process by using center pooling and cascade corner pooling.

\subsection{Object Detection as Keypoint Triplets}\label{Triplets}
The overall network architecture is shown in Figure~\ref{structure}. We represent each object by a center keypoint and a pair of corners. Specifically, we embed a heatmap for the center keypoints on the basis of CornerNet and predict the offsets of the center keypoints. Then, we use the method proposed in CornerNet~\cite{law2018cornernet} to generate top-$k$ bounding boxes. However, to effectively filter out the incorrect bounding boxes, we leverage the detected center keypoints and resort to the following procedure: (1) select top-$k$ center keypoints according to their scores; (2) use the corresponding offsets to remap these center keypoints to the input image; (3) define a central region for each bounding box and check if the central region contains center keypoints. Note that the class labels of the checked center keypoints should be same as that of the bounding box; (4) if a center keypoint is detected in the central region, we will preserve the bounding box. The score of the bounding box will be replaced by the average scores of the three points,~\ie,~the top-left corner, the bottom-right corner and the center keypoint. If there are no center keypoints detected in its central region, the bounding box will be removed.

\begin{figure}[tb]
  \centering 
  \subfigure[]{ 
    \includegraphics[width = 0.225\textwidth]{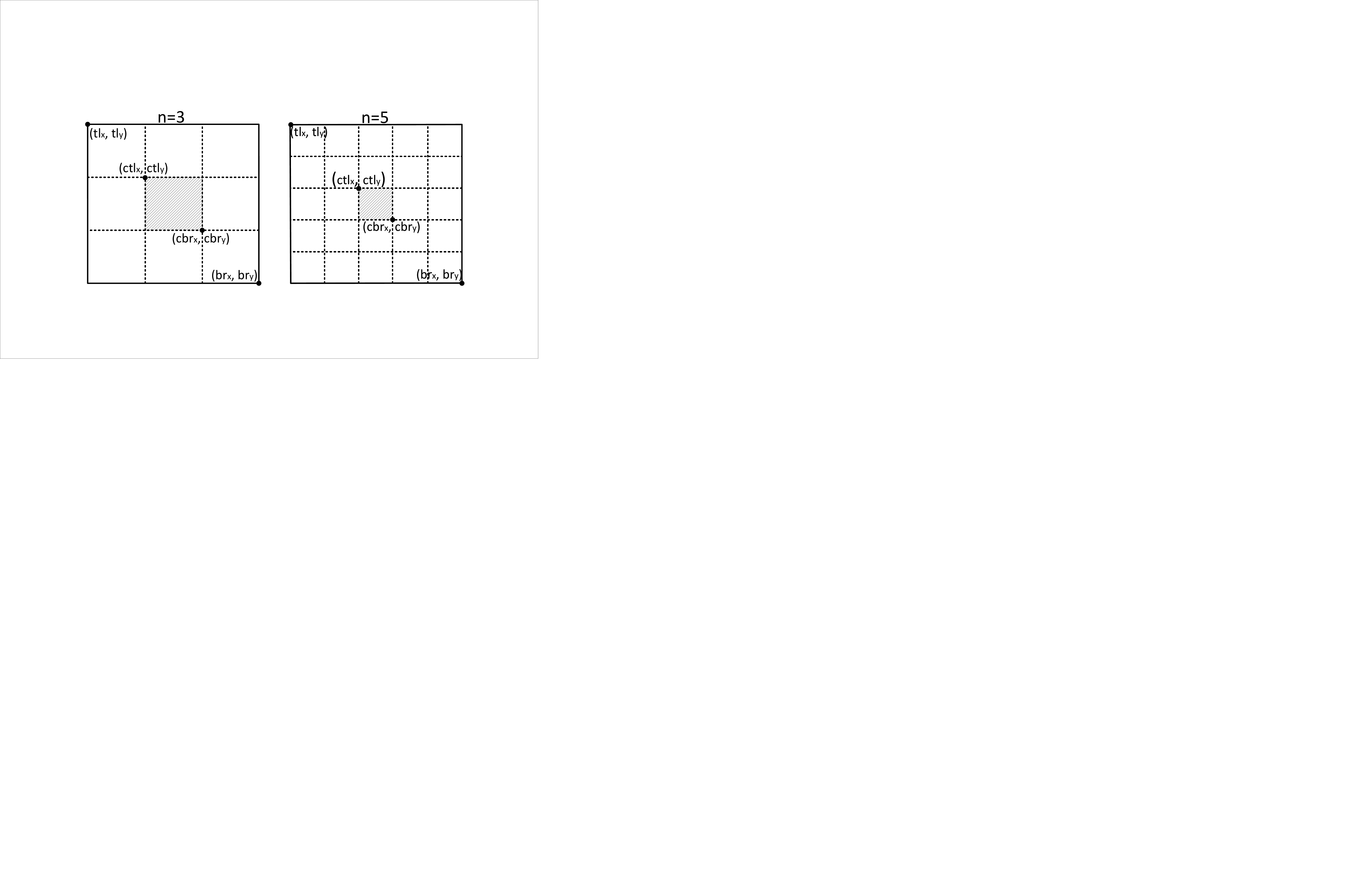}
    \label{fig_ncs1} 
  } 
  \subfigure[]{ 
    \includegraphics[width = 0.225\textwidth]{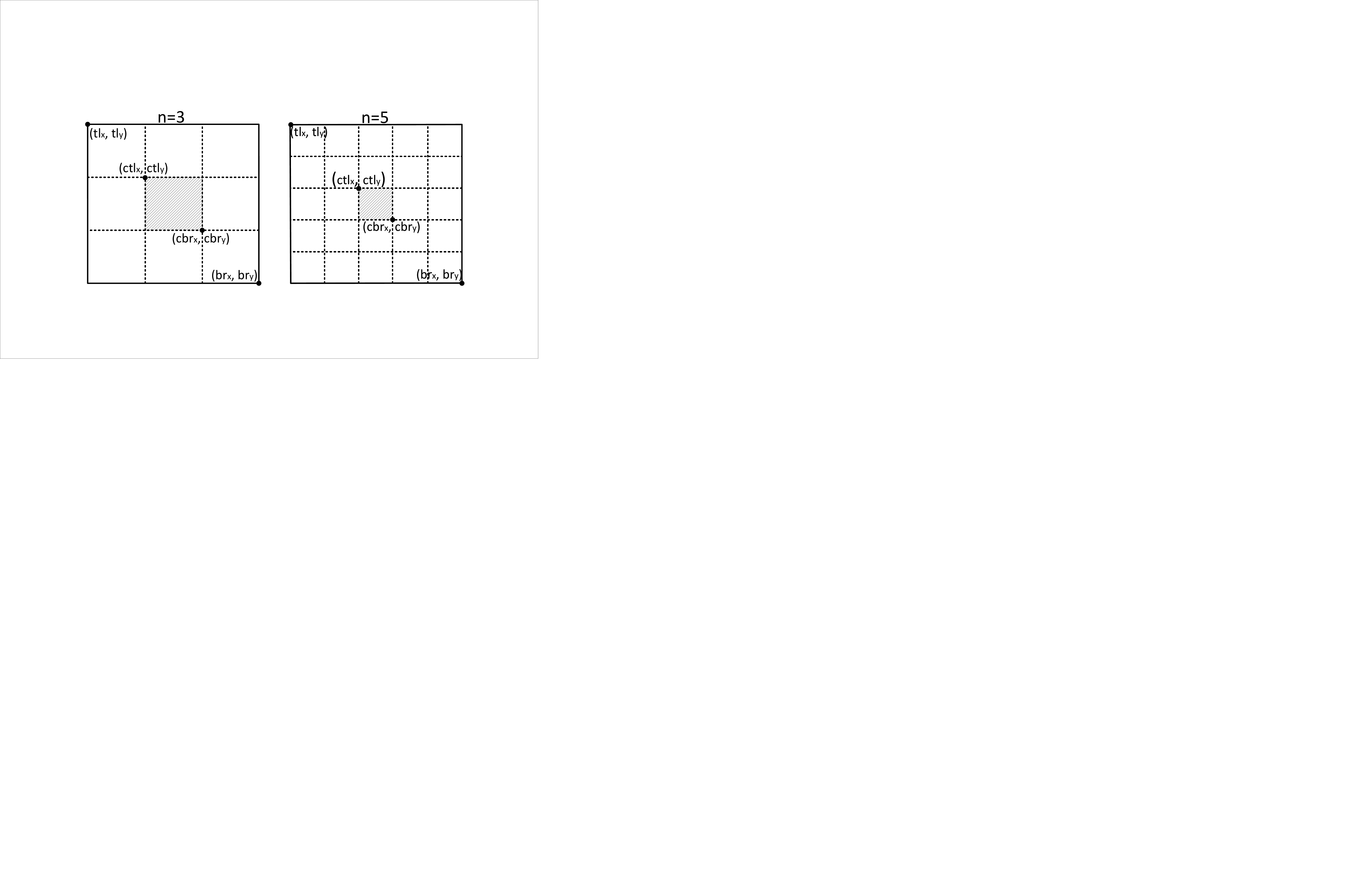}
    \label{fig_ncs2} 
  }
  \vspace{-2ex}
  \caption{(a) The central region when $n=3$. (b) The central region when $n=5$. The solid rectangles denote the predicted bounding boxes and the shaded regions denote the central regions.}
  \label{fig_ncs}
\end{figure}
The size of the central region in the bounding box affects the detection results. For example, smaller central regions lead to a low recall rate for small bounding boxes, while larger central regions lead to a low precision for large bounding boxes. Therefore, we propose a scale-aware central region to adaptively fit the size of bounding boxes. The scale-aware central region tends to generate a relatively large central region for a small bounding box, while a relatively small central region for a large bounding box. Suppose we want to determine if a bounding box $i$ needs to be preserved. Let $\mathrm{tl_{x}}$ and $\mathrm{tl_{y}}$ denote the coordinates of the top-left corner of $i$ and $\mathrm{br_{x}}$ and $\mathrm{br_{y}}$ denote the coordinates of the bottom-right corner of $i$. 
Define a central region $j$. Let $\mathrm{ctl_{x}}$ and $\mathrm{ctl_{y}}$ denote the coordinates of the top-left corner of $j$ and $\mathrm{cbr_{x}}$ and $\mathrm{cbr_{y}}$ denote the coordinates of the bottom-right corner of $j$. Then $\mathrm{tl_{x}}$, $\mathrm{tl_{y}}$, $\mathrm{br_{x}}$, $\mathrm{br_{y}}$, $\mathrm{ctl_{x}}$, $\mathrm{ctl_{y}}$, $\mathrm{cbr_{x}}$ and $\mathrm{cbr_{y}}$ should satisfy the following relationship:
\begin{equation} \label{eq_ncs1}
\small
\left\{
\begin{aligned}
&\mathrm{ctl_{x}} = \frac{(n+1)\mathrm{tl_{x}}+(n-1)\mathrm{br_{x}}}{2n} \\
&\mathrm{ctl_{y}} = \frac{(n+1)\mathrm{tl_{y}}+(n-1)\mathrm{br_{y}}}{2n} \\
&\mathrm{cbr_{x}} = \frac{(n-1)\mathrm{tl_{x}}+(n+1)\mathrm{br_{x}}}{2n} \\
&\mathrm{cbr_{y}} = \frac{(n-1)\mathrm{tl_{y}}+(n+1)\mathrm{br_{y}}}{2n} \\
\end{aligned}
\right.
\end{equation}
where $n$ is odd that determines the scale of the central region $j$. In this paper, $n$ is set to be $3$ and $5$ for the scales of bounding boxes less and greater than $150$, respectively. Figure~\ref{fig_ncs} shows two central regions when $n=3$ and $n=5$, respectively. According to Equation~(\ref{eq_ncs1}), we can determine a scale-aware central region, then we check if the central region contains center keypoints.

\subsection{Enriching Center and Corner Information}\label{Enriching}

\begin{figure}[tb]
  \centering 
  \subfigure[]{ 
    \includegraphics[width=0.11\textheight]{CenterPooling.pdf}
    \label{fig_ct}
  } 
  \subfigure[]{ 
    \includegraphics[width=0.11\textheight]{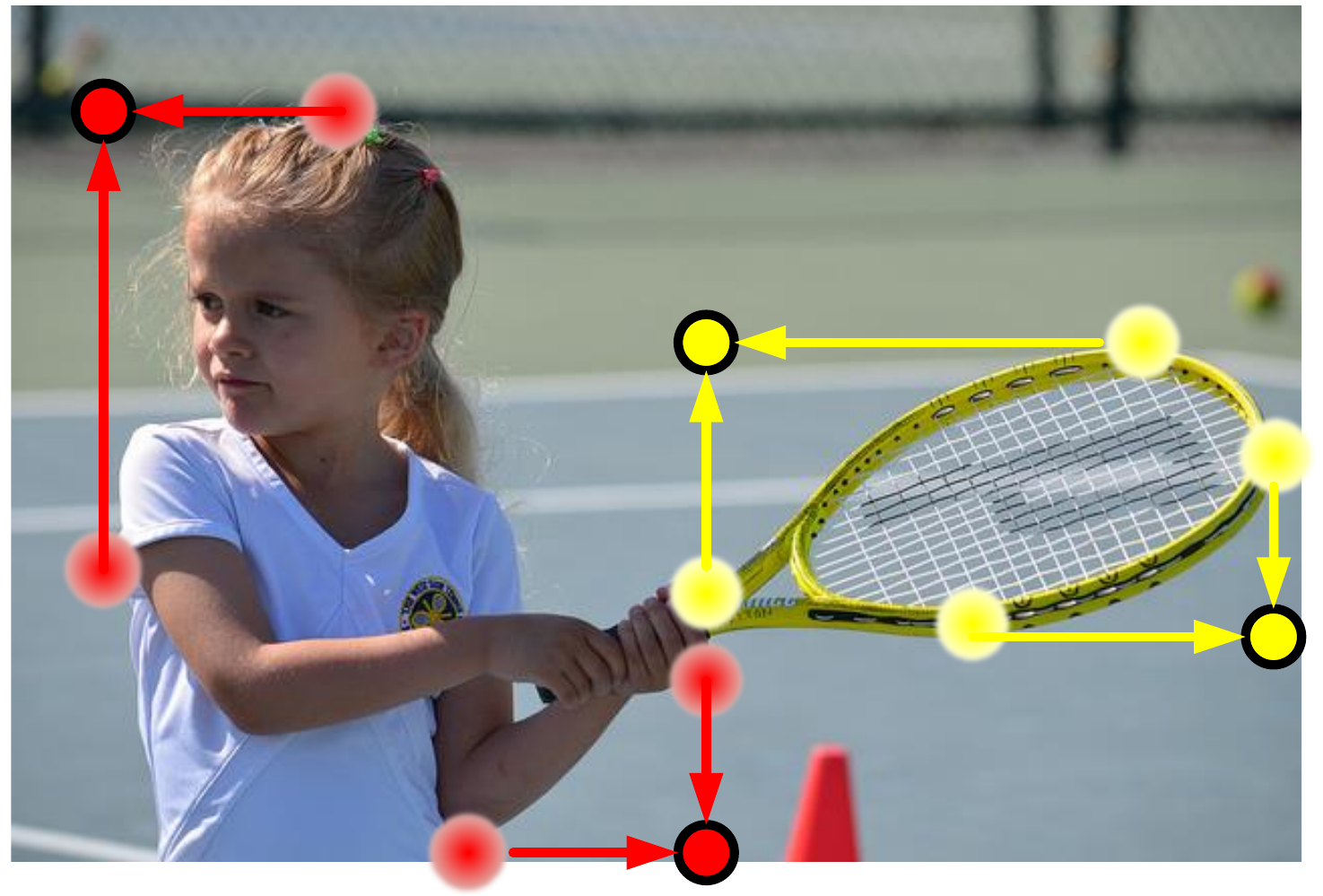}
    \label{fig_cp} 
  } 
  \subfigure[]{ 
    \includegraphics[width=0.11\textheight]{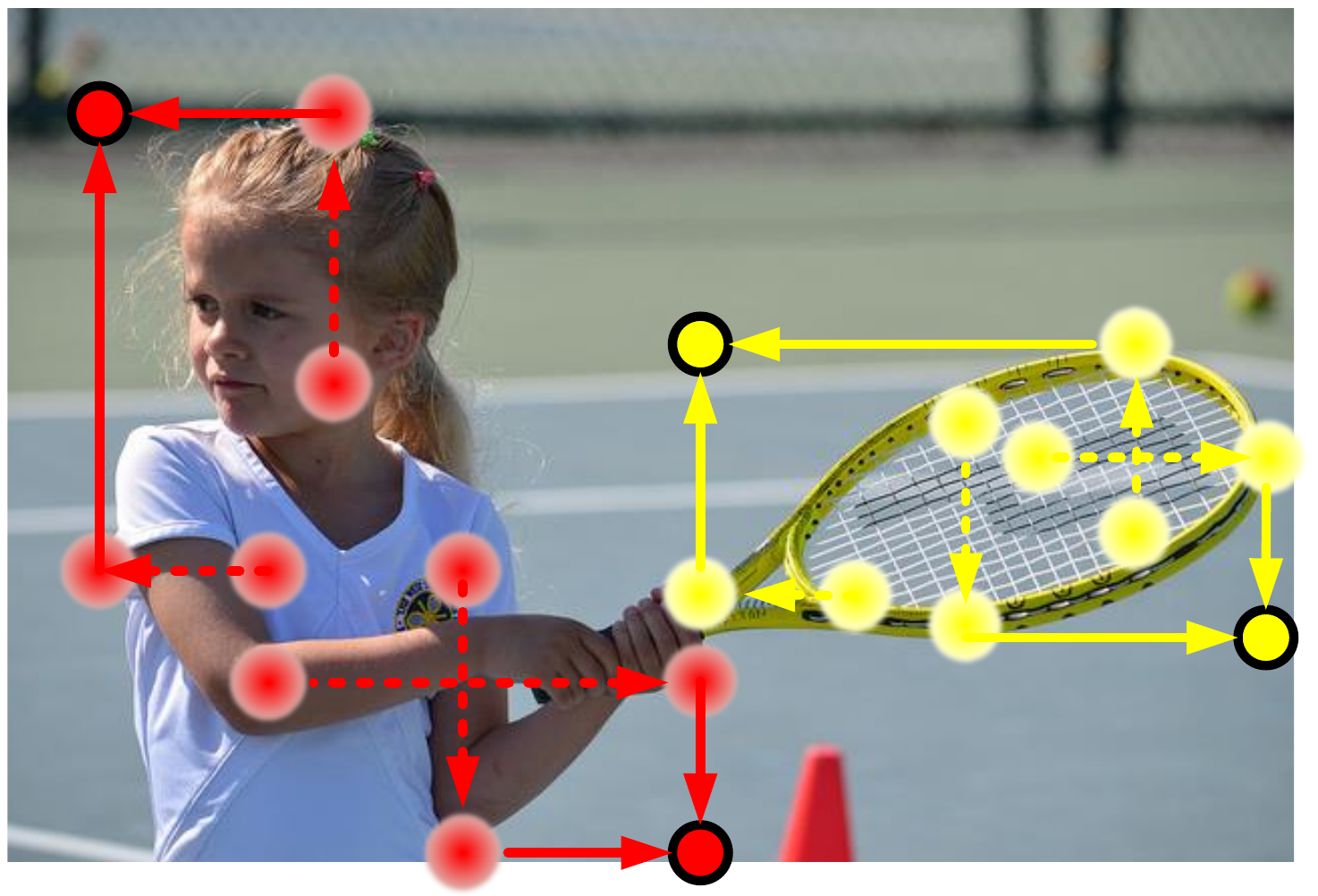}
    \label{fig_ccp} 
  } 
  \vspace{-2ex}
  \caption{(a) Center pooling takes the maximum values in both horizontal and vertical directions. (b) Corner pooling only takes the maximum values in boundary directions. (c) Cascade corner pooling takes the maximum values in both boundary directions and internal directions of objects.} 
  \label{fig_cornerpooling}
  \vspace{-3ex}
\end{figure}

\noindent\textbf{Center pooling.}~The geometric centers of objects do not necessarily convey very recognizable visual patterns (\eg, the human head contains strong visual patterns, but the center keypoint is often in the middle of the human body). To address this issue, we propose center pooling to capture richer and more recognizable visual patterns. Figure~\ref{fig_ct} shows the principle of center pooling. The detailed process of center pooling is as follows: the backbone outputs a feature map, and to determine if a pixel in the feature map is a center keypoint, we need to find the maximum value in its both horizontal and vertical directions and add them together. By doing this, center pooling helps the better detection of center keypoints.

\vspace{1ex}\noindent\textbf{Cascade corner pooling.}~Corners are often outside the objects, which lacks local appearance features. CornerNet~\cite{law2018cornernet} uses corner pooling to address this issue. The principle of corner pooling is shown in Figure~\ref{fig_cp}. Corner pooling aims to find the maximum values on the boundary directions so as to determine corners. However, it makes corners sensitive to the edges. To address this problem, we need to let corners ``see" the visual patterns of objects. The principle of cascade corner pooling is presented in Figure~\ref{fig_ccp}. It first looks along a boundary to find a boundary maximum value, then looks inside along the location of the boundary maximum value\footnote{For the topmost, leftmost, bottommost and rightmost boundary, look vertically towards the bottom, horizontally towards the right, vertically towards the top and horizontally towards the left, respectively.} to find an internal maximum value, and finally, add the two maximum values together. By doing this, the corners obtain both the the boundary information and the visual patterns of objects.

\begin{figure}[tb]
  \centering 
  \includegraphics[width=0.48\textwidth]{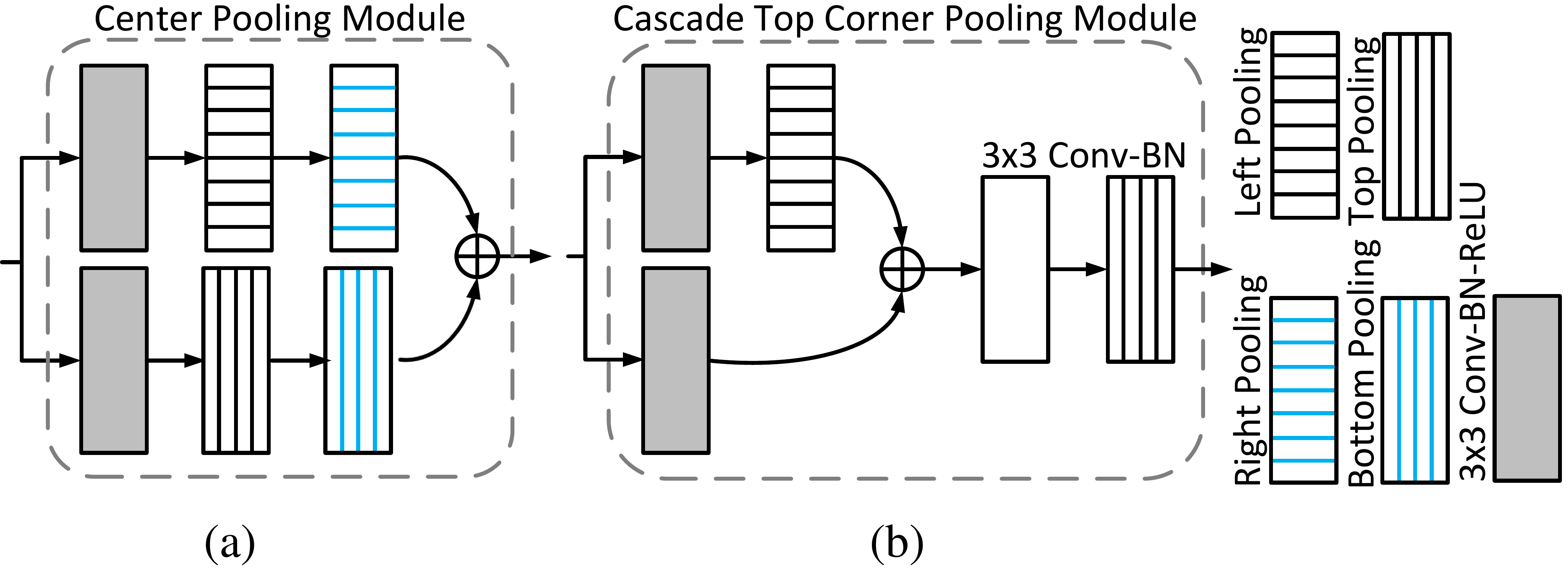}
  \vspace{-4ex}
  \caption{The structures of the center pooling module (a) and the cascade top corner pooling module (b). We achieve center pooling and the cascade corner pooling by combining the corner pooling at different directions.} 
  \label{PoolingStructure} 
\end{figure}

Both the center pooling and the cascade corner pooling can be easily achieved by combining the corner pooling~\cite{law2018cornernet} at different directions. Figure~\ref{PoolingStructure}{\color{red}(a)} shows the structure of the center pooling module. To take a maximum value in a direction, \eg, the horizontal direction, we only need to connect the left pooling and the right pooling in series. Figure~\ref{PoolingStructure}{\color{red}(b)} shows the structure of a cascade top corner pooling module. Compared with the top corner pooling in CornerNet~\cite{law2018cornernet}, we add a left corner pooling before the top corner pooling.

\subsection{Training and Inference }\label{Training}
\noindent\textbf{Training.} Our method is implemented in Pytorch~\cite{paszke2017automatic} and the network is trained from scratch. The resolution of the input image is $511\times511$, leading to heatmaps of size $128\times128$. We use the data augmentation strategy presented in~\cite{law2018cornernet} to train a robust model. Adam~\cite{kingma2014adam} is used to optimize the training loss:
\begin{equation} \label{loss} 
\small
\mathrm{L = L_{det}^\mathrm{co} + L_{det}^\mathrm{ce} + \alpha L_{pull}^\mathrm{co} + \beta L_{push}^\mathrm{co} + \gamma\left( L_{off}^\mathrm{co} + L_{off}^\mathrm{ce} \right)},
\end{equation}
where $\mathrm{L_{det}^\mathrm{co}}$ and $\mathrm{L_{det}^\mathrm{ce}}$ denote the focal losses, which are used to train the network to detect corners and center keypoints, respectively. $\mathrm{L_{pull}^\mathrm{co}}$ is a ``pull'' loss for corners, which is used to minimize the distance of the embedding vectors that belongs to the same objects. $\mathrm{L_{push}^\mathrm{co}}$ is a  ``push'' loss for corners, which is used to maximize the distance of the embedding vectors that belongs to different objects. $\mathrm{L_{off}^\mathrm{co}}$ and $\mathrm{L_{off}^\mathrm{ce}}$ are $\ell_1$-losses~\cite{girshick2015fast}, which are used to train the network to predict the offsets of corners and center keypoints, respectively. $\mathrm{\alpha}$, $\mathrm{\beta}$ and $\mathrm{\gamma}$ denote the weights for corresponding losses, which are set to $0.1$, $0.1$ and $1$, respectively. $\mathrm{L_{det}}$, $\mathrm{L_{pull}}$, $\mathrm{L_{push}}$ and $\mathrm{L_{off}}$ are all defined in the CornerNet, we suggest to refer to~\cite{law2018cornernet} for details. We train the CenterNet on $8$ Tesla V100 (32GB) GPUs and use a batch size of $48$. The maximum number of iterations is $480\mathrm{K}$. We use a learning rate of $2.5\times10^{-4}$ for the first $450\mathrm{K}$ iterations and then continue training $30\mathrm{K}$ iterations with a rate of $2.5\times10^{-5}$. 

\vspace{1ex}\noindent\textbf{Inference.} Following~\cite{law2018cornernet}, for the single-scale testing, we input both the original and horizontally flipped images with the original resolutions into the network. While for the multi-scale testing, we input both the original and horizontally flipped images with the resolutions of $0.6, 1, 1.2, 1.5$ and $1.8$. We select top $70$ center keypoints, top $70$ top-left corners and top $70$ bottom-right corners from the heatmaps to detect the bounding boxes. We flip the bounding boxes detected in the horizontally flipped images and mix them into the original bounding boxes. Soft-nms~\cite{bodla2017soft} is used to remove the redundant bounding boxes. We finally select top $100$ bounding boxes according to their scores as the final detection results.
\section{Experiments}
\label{Experiments}

\subsection{Dataset, Metrics and Baseline}
\label{sec:setting}

We evaluate our method on the MS-COCO dataset~\cite{lin2014microsoft}. It contains $80$ categories and more than $1.5$ million object instances. The large number of small objects makes it a very challenging dataset. We use the `trainval35k' set~\cite{hoiem2012diagnosing} (\ie, $80\mathrm{K}$ training images and $35\mathrm{K}$ validation images) for training and test the results on the test-dev set. We use another $5\mathrm{K}$ images in the validation set to perform ablation studies and visualization experiments.

MS-COCO dataset~\cite{lin2014microsoft} uses AP and AR metrics to characterize the performance of a detector. AP represents the average precision rate, which is computed over ten different IoU thresholds (\ie, $0.5:0.05:0.95$) and all categories. It is considered the single most important metric on the MS-COCO dataset. AR represents the maximum recall rate, which is computed over a fixed number of detections (\ie, $1$, $10$ and $100$ ) per image and averaged over all categories and the ten different IoU thresholds. Additionally, AP and AR can be used to evaluate the performance under different object scales, including small objects ($\mathrm{area}<32^{2}$), medium objects ($32^{2}<\mathrm{area}<96^{2}$) and large objects ($\mathrm{area}>96^{2}$).

Our direct baseline is CornerNet~\cite{law2018cornernet}. Following it, we use the stacked hourglass network~\cite{newell2016stacked} with $52$ and $104$ layers as the backbone -- the latter has two hourglass modules while the former has only one. All modifications on the hourglass architecture, made by~\cite{law2018cornernet}, are preserved.

\begin{table*}[tb]
\small
\centering
\resizebox{1.0\textwidth}{!}{
\begin{tabular}{|l|l|cc|cccccc|cccccc|}
\hline
Method & Backbone & Train input & Test input& AP & AP$_{50}$ & AP$_{75}$ & AP$_\mathrm{S}$ & AP$_\mathrm{M}$ & AP$_\mathrm{L}$ & AR$_1$ & AR$_{10}$ & AR$_{100}$ & AR$_\mathrm{S}$ & AR$_\mathrm{M}$ & AR$_\mathrm{L}$\\
\hline
\hline
\textbf{Two-stage:} & & & & & & & & & & & & & & &\\
DeNet~\cite{tychsen2017denet} & ResNet-101~\cite{he2016deep} & 512$\times$512 & 512$\times$512 & 33.8 & 53.4 & 36.1 & 12.3 & 36.1 & 50.8 & 29.6 & 42.6 & 43.5 & 19.2 & 46.9 & 64.3 \\
CoupleNet~\cite{zhu2017couplenet} & ResNet-101 & ori.& ori. & 34.4 & 54.8 & 37.2 & 13.4 & 38.1 & 50.8 & 30.0 & 45.0 & 46.4 & 20.7 & 53.1 & 68.5 \\
Faster R-CNN by G-RMI~\cite{huang2017speed} & Inception-ResNet-v2~\cite{szegedy2017inception} & $\sim$ 1000$\times$600 & $\sim$ 1000$\times$600 & 34.7 & 55.5 & 36.7 & 13.5 & 38.1 & 52.0 & - & - & - & - & - & - \\
Faster R-CNN +++~\cite{he2016deep} & ResNet-101 & $\sim$ 1000$\times$600 & $\sim$ 1000$\times$600 & 34.9 & 55.7 & 37.4 & 15.6 & 38.7 & 50.9 & - & - & - & - & - & - \\
Faster R-CNN w/ FPN~\cite{lin2017feature} & ResNet-101 & $\sim$ 1000$\times$600 & $\sim$ 1000$\times$600 & 36.2 & 59.1 & 39.0 & 18.2 & 39.0 & 48.2 & - & - & - & - & - & - \\
Faster R-CNN w/ TDM~\cite{shrivastava2016beyond} & Inception-ResNet-v2 &  - & - & 36.8 & 57.7 & 39.2 & 16.2 & 39.8 & 52.1 & \textbf{31.6} & \textbf{49.3} & \textbf{51.9} & \textbf{28.1} & \textbf{56.6} & \textbf{71.1} \\
D-FCN~\cite{dai2017deformable} & Aligned-Inception-ResNet & $\sim$ 1000$\times$600 & $\sim$ 1000$\times$600 & 37.5 & 58.0 & - & 19.4 & 40.1 & 52.5 & - & - & - & - & - & - \\
Regionlets~\cite{xu2018deep} & ResNet-101 & $\sim$ 1000$\times$600 & $\sim$ 1000$\times$600 & 39.3 & 59.8 & - & 21.7 & 43.7 & 50.9 & - & - & - & - & - & - \\
Mask R-CNN~\cite{he2017mask} & ResNeXt-101 & $\sim$ 1300$\times$800 & $\sim$ 1300$\times$800 & 39.8 & 62.3 & 43.4 & 22.1 & 43.2 & 51.2 & - & - & - & - & - & - \\
Soft-NMS~\cite{bodla2017soft} & Aligned-Inception-ResNet & $\sim$ 1300$\times$800 & $\sim$ 1300$\times$800 & 40.9 & 62.8 & - & 23.3 & 43.6 & 53.3 & - & - & - & - & - & - \\
Fitness R-CNN~\cite{tychsen2018improving} & ResNet-101 & 512$\times$512 & 1024$\times$1024 & 41.8 & 60.9 & 44.9 & 21.5 & 45.0 & 57.5 & - & - & - & - & - & - \\
Cascade R-CNN~\cite{cai2018cascade} & ResNet-101 & - & - & 42.8 & 62.1 & 46.3 & 23.7 & 45.5 & 55.2 & - & - & - & - & - & - \\
Grid R-CNN w/ FPN~\cite{Lu2018Grid} & ResNeXt-101 & $\sim$ 1300$\times$800 & $\sim$ 1300$\times$800 & 43.2 & 63.0 & 46.6 & 25.1 & 46.5 & 55.2 & - & - & - & - & - & - \\
D-RFCN + SNIP (multi-scale)~\cite{singh2018analysis} & DPN-98~\cite{chen2017dual} & $\sim$ 2000$\times$1200 & $\sim$ 2000$\times$1200 & 45.7 & \textbf{67.3} & 51.1 & 29.3 & 48.8 & 57.1 & - & - & - & - & - & - \\
PANet (multi-scale)~\cite{liu2018path} & ResNeXt-101 & $\sim$ 1400$\times$840 & $\sim$ 1400$\times$840 & \textbf{47.4} & 67.2 & \textbf{51.8} & \textbf{30.1} & \textbf{51.7} & \textbf{60.0} & - & - & - & - & - & - \\
\hline
\hline
\textbf{One-stage:} & & & & & & & & & & & & & & &\\
YOLOv2~\cite{redmon2017yolo9000} & DarkNet-19 & 544$\times$544 & 544$\times$544 & 21.6 & 44.0 & 19.2 & 5.0 & 22.4 & 35.5 & 20.7 & 31.6 & 33.3 & 9.8 & 36.5 & 54.4 \\
DSOD300~\cite{shen2017dsod} & DS/64-192-48-1 & 300$\times$300 & 300$\times$300 & 29.3 & 47.3 & 30.6 & 9.4 & 31.5 & 47.0 & 27.3 & 40.7 & 43.0 & 16.7 & 47.1 & 65.0 \\
GRP-DSOD320~\cite{shen2017learning} & DS/64-192-48-1 & 320$\times$320 & 320$\times$320 & 30.0 & 47.9 & 31.8 & 10.9 & 33.6 & 46.3 & 28.0 & 42.1 & 44.5 & 18.8 & 49.1 & 65.0 \\
SSD513~\cite{liu2016ssd} & ResNet-101 & 513$\times$513 & 513$\times$513 & 31.2 & 50.4 & 33.3 & 10.2 & 34.5 & 49.8 & 28.3 & 42.1 & 44.4 & 17.6 & 49.2 & 65.8 \\
DSSD513~\cite{fu2017dssd} & ResNet-101 & 513$\times$513 & 513$\times$513 & 33.2 & 53.3 & 35.2 & 13.0 & 35.4 & 51.1 & 28.9 & 43.5 & 46.2 & 21.8 & 49.1 & 66.4 \\
RefineDet512 (single-scale)~\cite{zhang2018single} & ResNet-101 & 512$\times$512 & 512$\times$512 & 36.4 & 57.5 & 39.5 & 16.6 & 39.9 & 51.4 & - & - & - & - & - & - \\
CornerNet511 (single-scale)~\cite{law2018cornernet} & Hourglass-52 & 511$\times$511 & ori. & 37.8 & 53.7 & 40.1 & 17.0 & 39.0 & 50.5 & 33.9 & 52.3 & 57.0 & 35.0 & 59.3 & 74.7 \\
RetinaNet800~\cite{lin2017focal} & ResNet-101 & 800$\times$800 & 800$\times$800 & 39.1 & 59.1 & 42.3 & 21.8 & 42.7 & 50.2 & - & - & - & - & - & - \\
CornerNet511 (multi-scale)~\cite{law2018cornernet} & Hourglass-52 & 511$\times$511 & $\le$1.5$\times$ & 39.4 & 54.9 & 42.3 & 18.9 & 41.2 & 52.7 & 35.0 & 53.5 & 57.7 & 36.1 & 60.1 & 75.1 \\
CornerNet511 (single-scale)~\cite{law2018cornernet} & Hourglass-104 & 511$\times$511 & ori. & 40.5 & 56.5 & 43.1 & 19.4 & 42.7 & 53.9 & 35.3 & 54.3 & 59.1 & 37.4 & 61.9 & 76.9 \\
RefineDet512 (multi-scale)~\cite{zhang2018single} & ResNet-101 & 512$\times$512 & $\le$2.25$\times$ & 41.8 & 62.9 & 45.7 & 25.6 & 45.1 & 54.1 &  &  &  &  &  &  \\
CornerNet511 (multi-scale)~\cite{law2018cornernet} & Hourglass-104 & 511$\times$511 & $\le$1.5$\times$ & 42.1 & 57.8 & 45.3 & 20.8 & 44.8 & 56.7 & 36.4 & 55.7 & 60.0 & 38.5 & 62.7 & 77.4 \\
\hline
\textbf{CenterNet511} (single-scale) & Hourglass-52 & 511$\times$511 & ori. & \mycolor{41.6} & \mycolor{59.4} & \mycolor{44.2} & \mycolor{22.5} & \mycolor{43.1} & \mycolor{54.1} & \mycolor{34.8} & \mycolor{55.7} & \mycolor{60.1} & \mycolor{38.6} & \mycolor{63.3} & \mycolor{76.9} \\
\textbf{CenterNet511} (single-scale) & Hourglass-104 & 511$\times$511 & ori. & \mycolor{44.9} & \mycolor{62.4} & \mycolor{48.1} & \mycolor{25.6} & \mycolor{47.4} & \mycolor{57.4} & \mycolor{36.1} & \mycolor{58.4} & \mycolor{63.3} & \mycolor{41.3} & \mycolor{67.1} & \mycolor{80.2} \\
\textbf{CenterNet511} (multi-scale) & Hourglass-52 & 511$\times$511 & $\le$1.8$\times$ & \mycolor{43.5} & \mycolor{61.3} & \mycolor{46.7} & \mycolor{25.3} & \mycolor{45.3} & \mycolor{55.0} & \mycolor{36.0} & \mycolor{57.2} & \mycolor{61.3} & \mycolor{41.4} & \mycolor{64.0} & \mycolor{76.3} \\
\textbf{CenterNet511} (multi-scale) & Hourglass-104 &  511$\times$511 & $\le$1.8$\times$ & \mycolor{\textbf{47.0}} & \mycolor{\textbf{64.5}} & \mycolor{\textbf{50.7}} & \mycolor{\textbf{28.9}} & \mycolor{\textbf{49.9}} & \mycolor{\textbf{58.9}} & \mycolor{\textbf{37.5}} & \mycolor{\textbf{60.3}} & \mycolor{\textbf{64.8}} & \mycolor{\textbf{45.1}} & \mycolor{\textbf{68.3}} & \mycolor{\textbf{79.7}} \\
\hline
\end{tabular}}
\vspace{-2ex}
\caption{Performance comparison ($\%$) with the state-of-the-art methods on the MS-COCO test-dev dataset. CenterNet outperforms all existing one-stage detectors by a large margin and ranks among the top of state-of-the-art two-stage detectors.}
\label{tab1}
\vspace{-2ex}
\end{table*}

\subsection{Comparisons with State-of-the-art Detectors}

Table~\ref{tab1} shows the comparison with the state-of-the-art detectors on the MS-COCO test-dev set.

Compared with the baseline CornerNet~\cite{law2018cornernet}, the proposed CenterNet achieves a remarkable improvement. For example, CenterNet511-52 (means that the resolution of input images is $511\times 511$ and the backbone is Hourglass-52) reports a single-scale testing AP of $41.6\%$, an improvement of $3.8\%$ over $37.8\%$, and a multi-scale testing AP of $43.5\%$, an improvement of $4.1\%$ over $39.4\%$, achieved by CornerNet under the same setting. When using the deeper backbone (\ie, Hourglass-104), the AP improvement over CornerNet are $4.4\%$ (from $40.5\%$ to $44.9\%$) and $4.9\%$ (from $42.1\%$ to $47.0\%$) under the single-scale and multi-scale testing, respectively. These results firmly demonstrate the effectiveness of CenterNet.

\begin{figure*}[t]
  \subfigure{ 
    \includegraphics[height=0.122\textwidth,width=0.07\textheight]{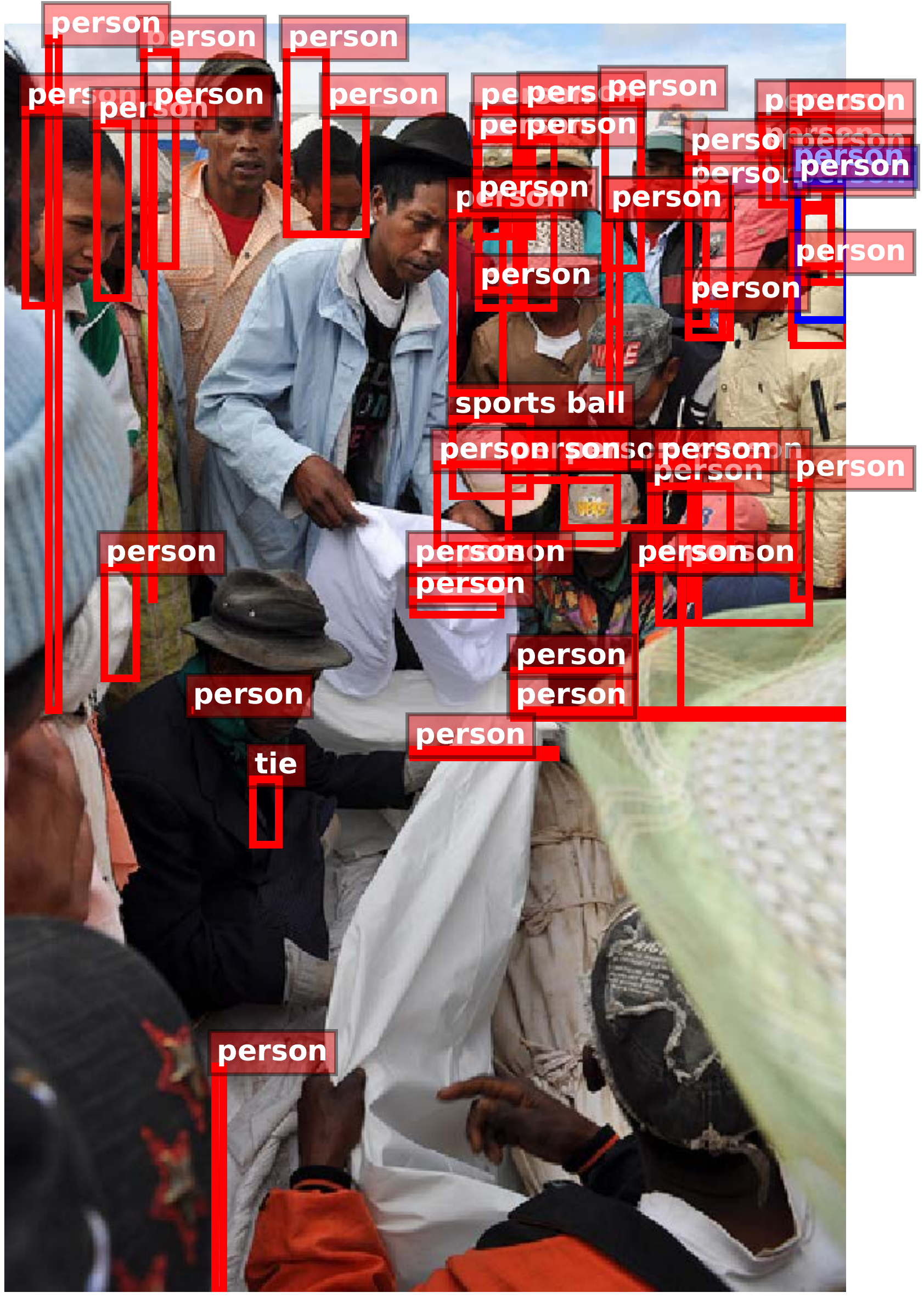}
    \label{fig7_11}
  } 
  \hspace{-0.1in}
  \subfigure{ 
    \includegraphics[height=0.12\textwidth,width=0.13\textheight]{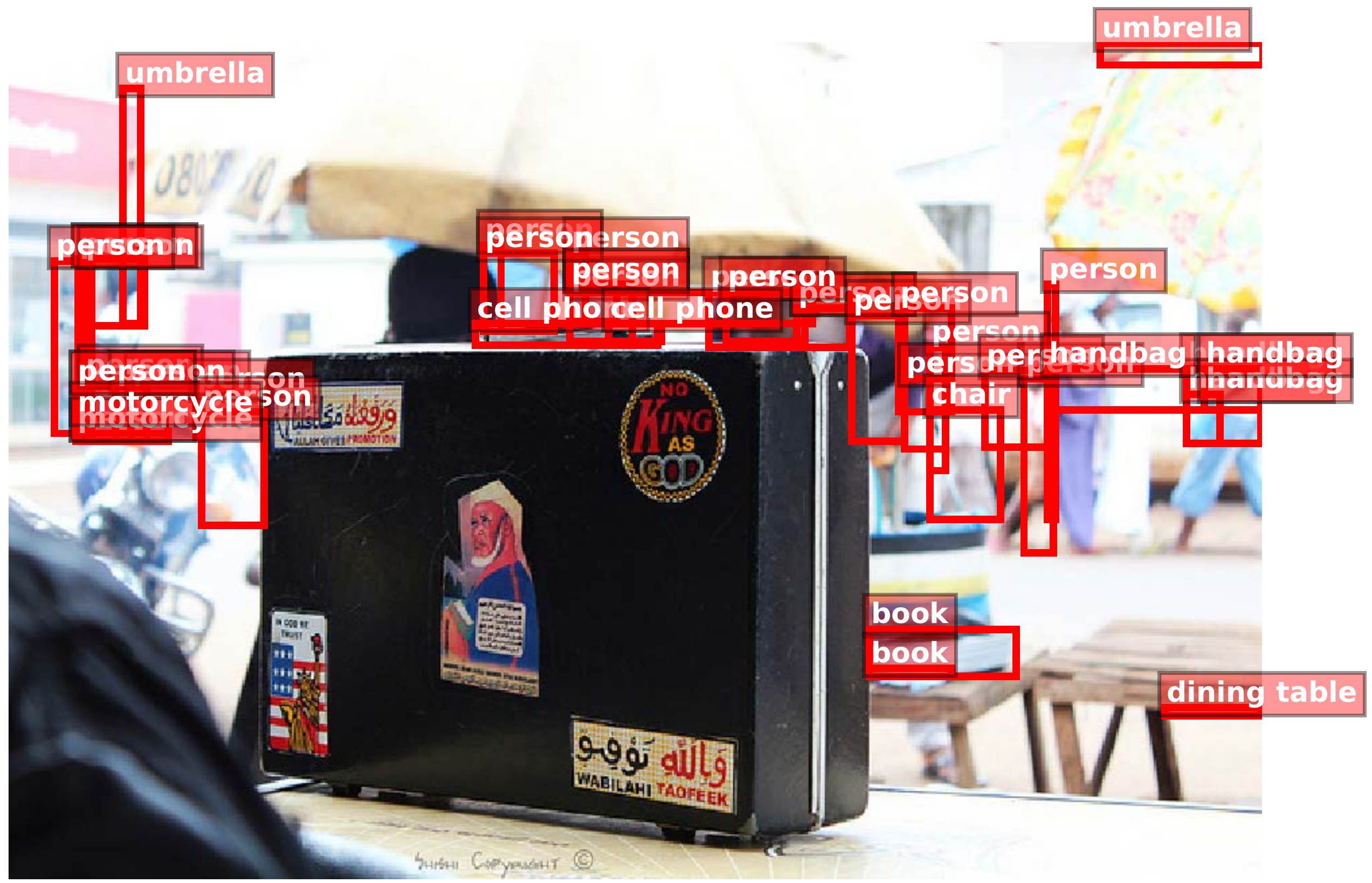}
    \label{fig7_21}
  } 
  \hspace{-0.11in}
  \subfigure{  
    \includegraphics[height=0.125\textwidth,width=0.14\textheight]{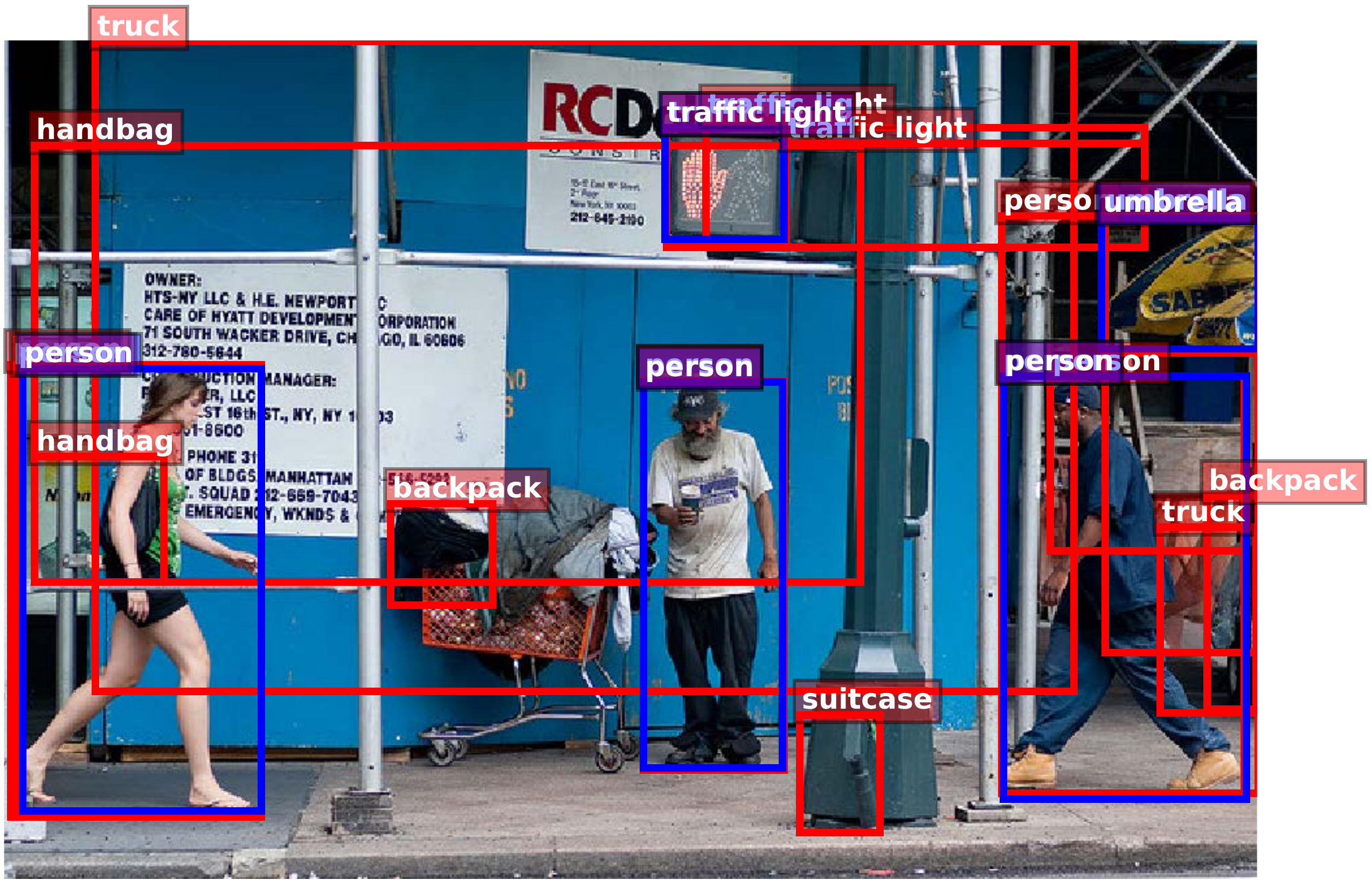}
    \label{fig7_31}
  } 
  \hspace{-0.11in}
  \subfigure{ 
    \includegraphics[height=0.12\textwidth,width=0.13\textheight]{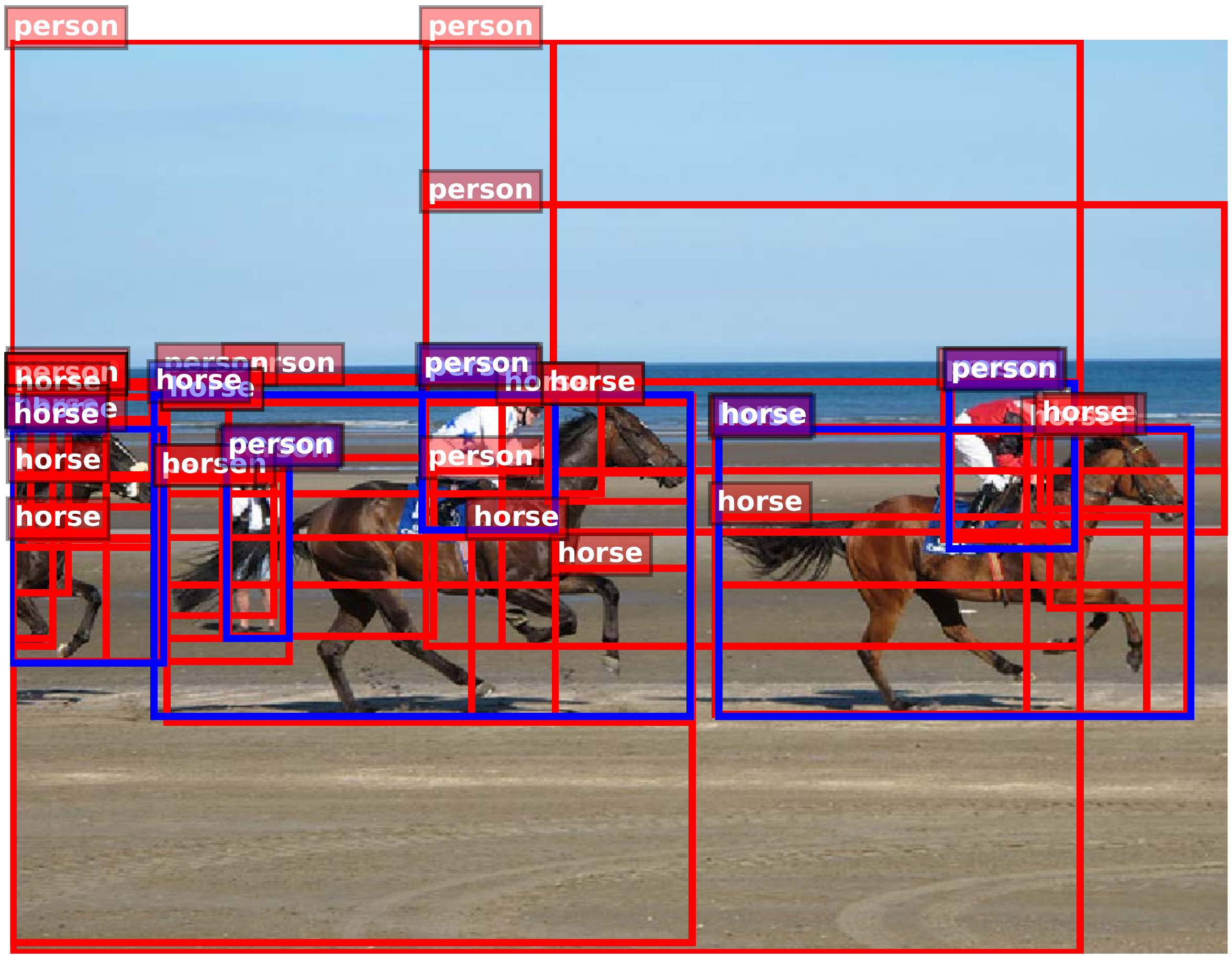}
    \label{fig7_41}
  } 
  \hspace{-0.11in}
  \subfigure{ 
    \includegraphics[height=0.12\textwidth,width=0.13\textheight]{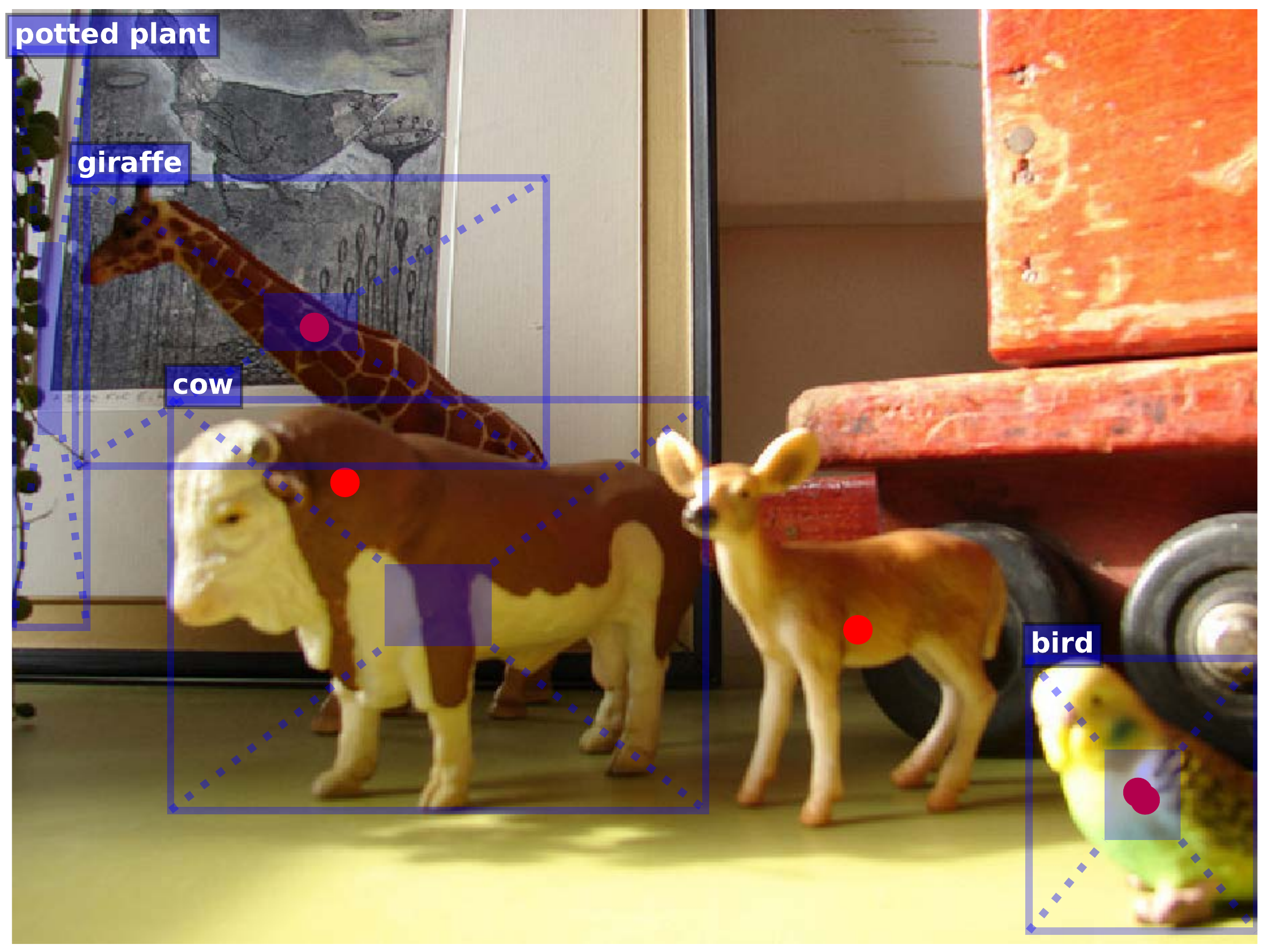}
    \label{fig7_51}
  }
  \vspace{-0.12in}
  \hspace{-0.11in}
  \subfigure{ 
    \includegraphics[height=0.12\textwidth,width=0.14\textheight]{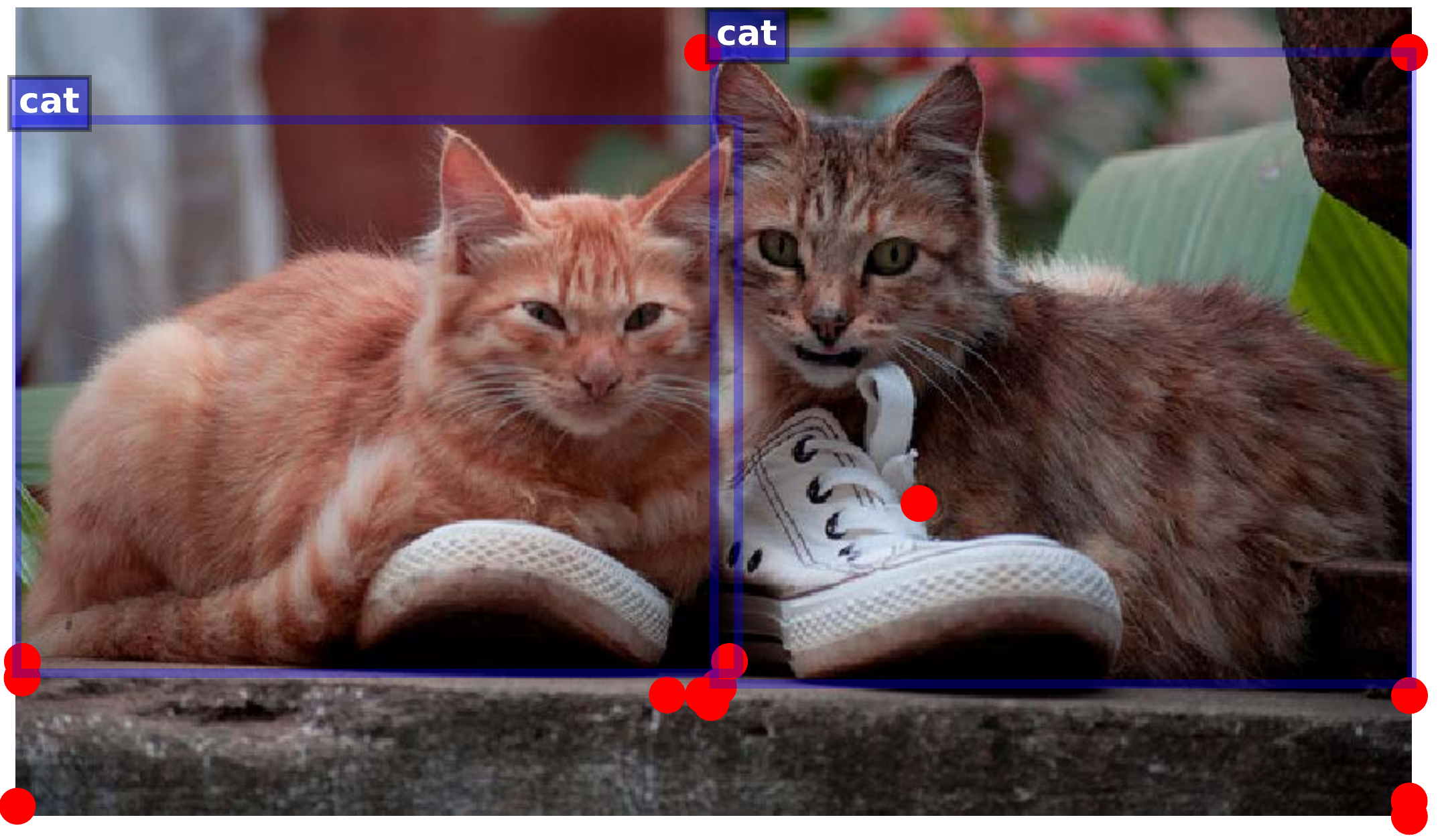}
    \label{fig7_61}
  }
  \renewcommand\thesubfigure{(a)}
  \hspace{-0.1in}
  \subfigure[]{ 
    \includegraphics[height=0.122\textwidth,width=0.071\textheight]{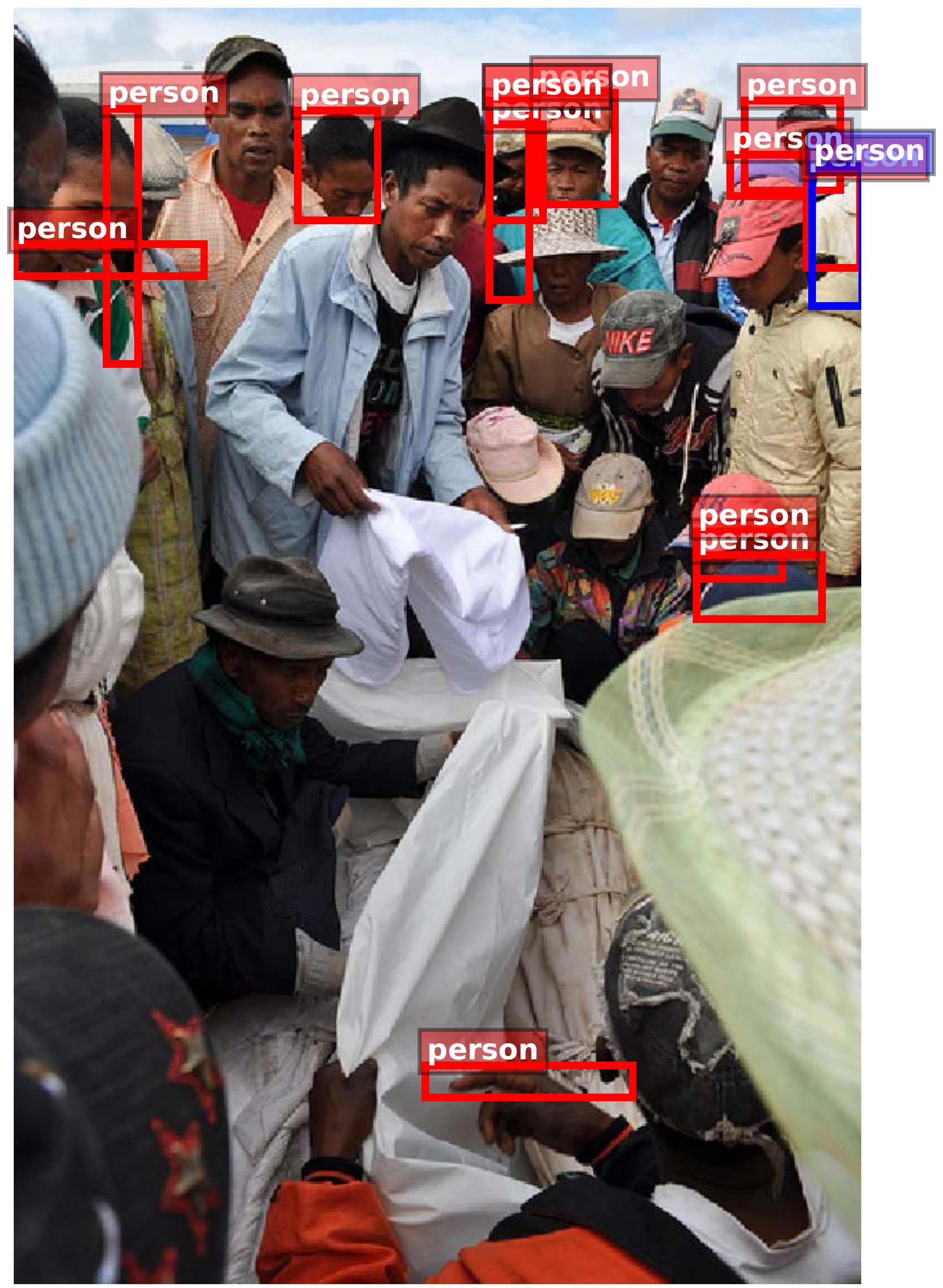}
    \label{fig7_12}
  } 
  \renewcommand\thesubfigure{(b)}
  \hspace{-0.11in}
  \subfigure[]{ 
    \includegraphics[height=0.12\textwidth,width=0.122\textheight]{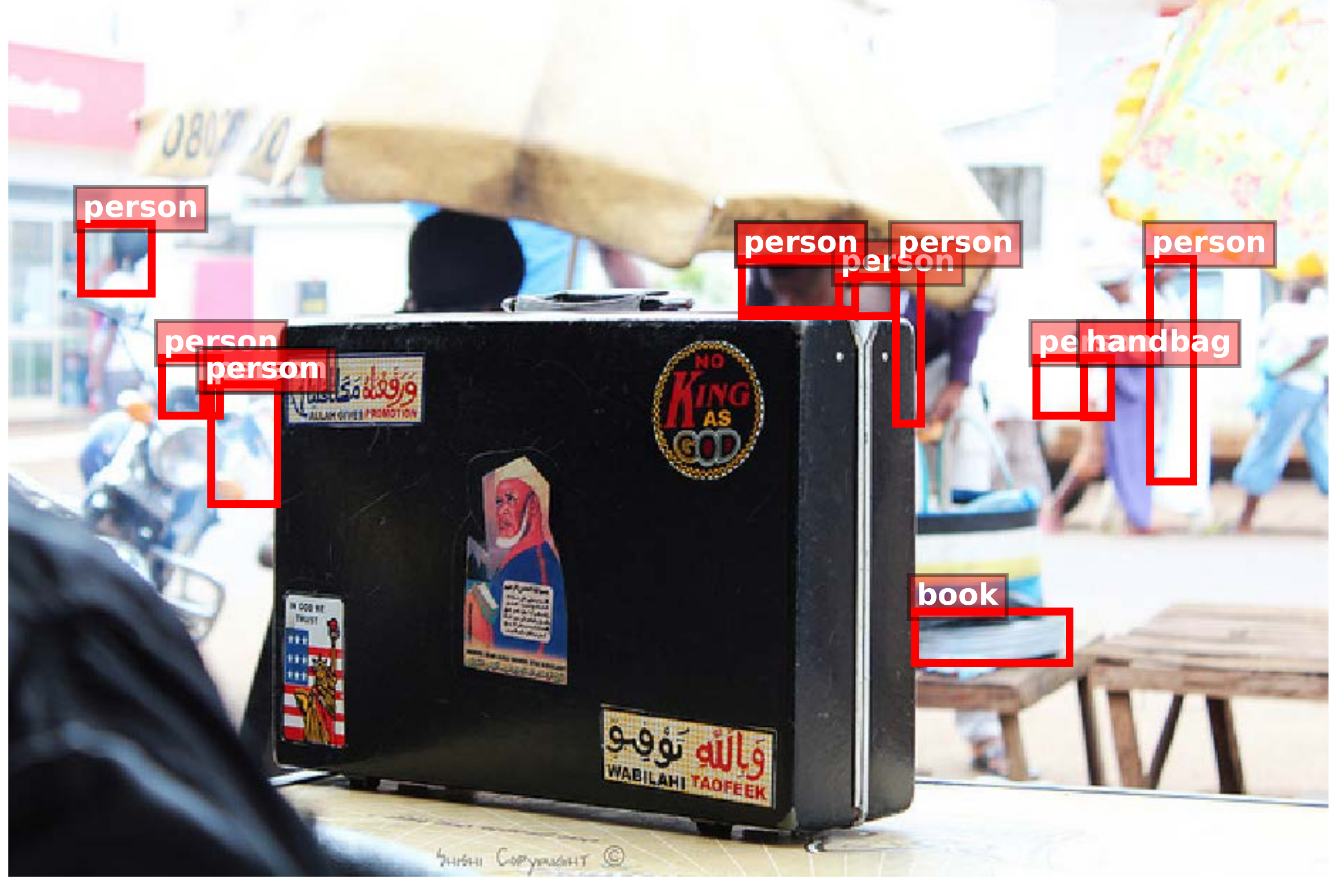}
    \label{fig7_22}
  } 
  \renewcommand\thesubfigure{(c)}
  \hspace{-0.04in}
  \subfigure[]{ 
    \includegraphics[height=0.12\textwidth,width=0.13\textheight]{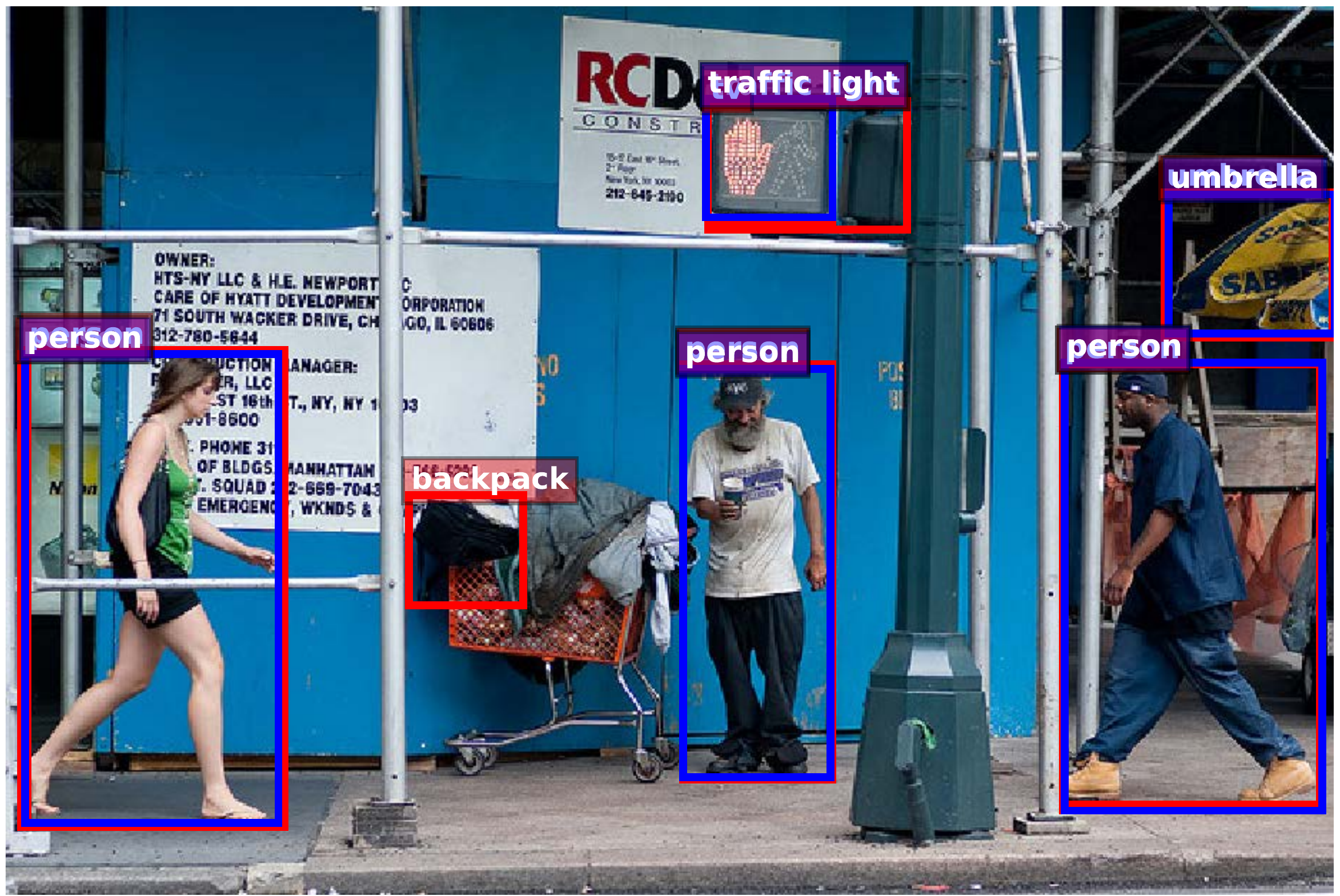}
    \label{fig7_32}
  } 
  \renewcommand\thesubfigure{(d)}
  \hspace{-0.02in}
  \subfigure[]{ 
    \includegraphics[height=0.12\textwidth,width=0.13\textheight]{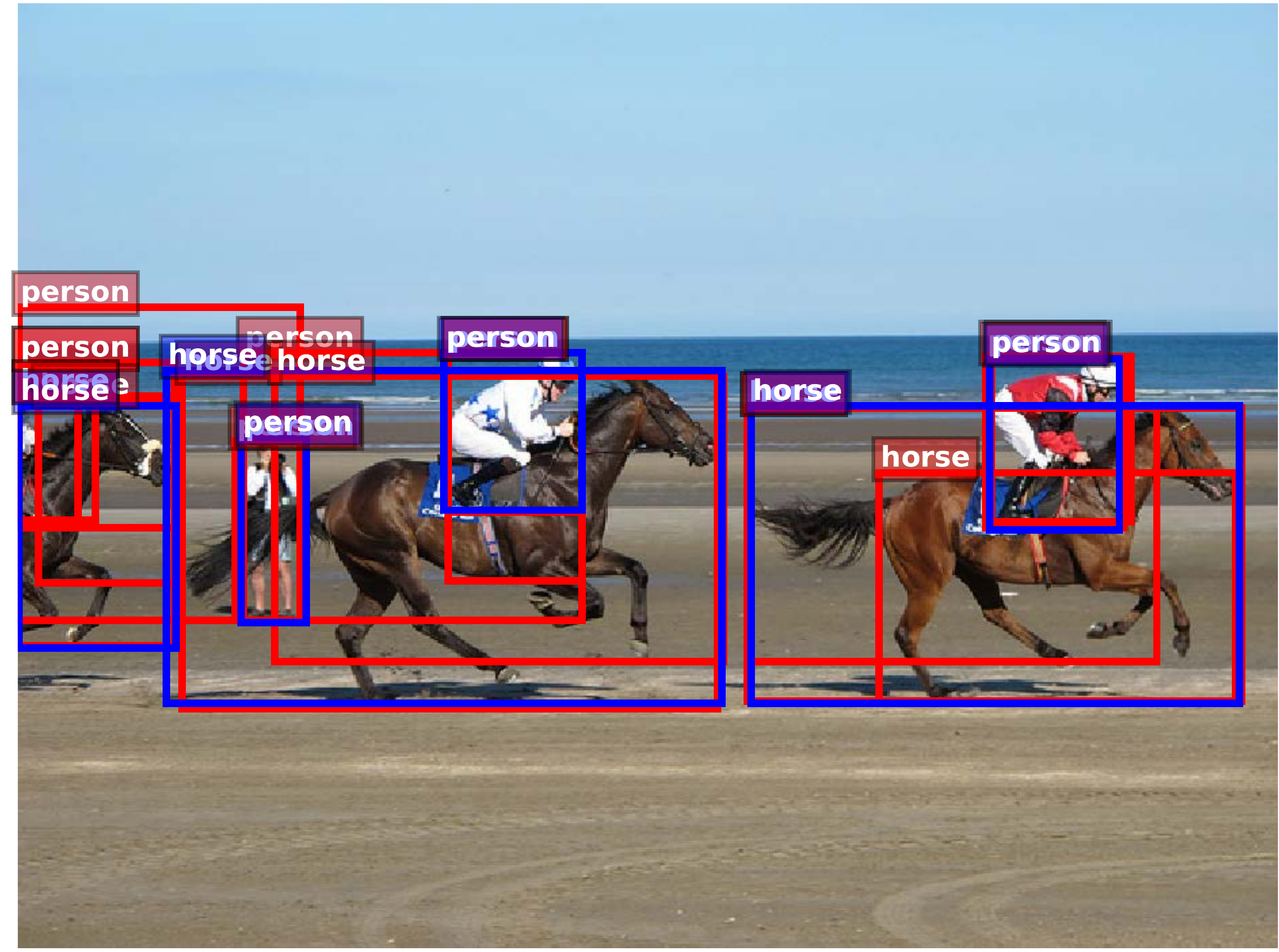}
    \label{fig7_42}
  }
  \renewcommand\thesubfigure{(e)}
  \hspace{-0.1in}
  \subfigure[]{ 
    \includegraphics[height=0.12\textwidth,width=0.13\textheight]{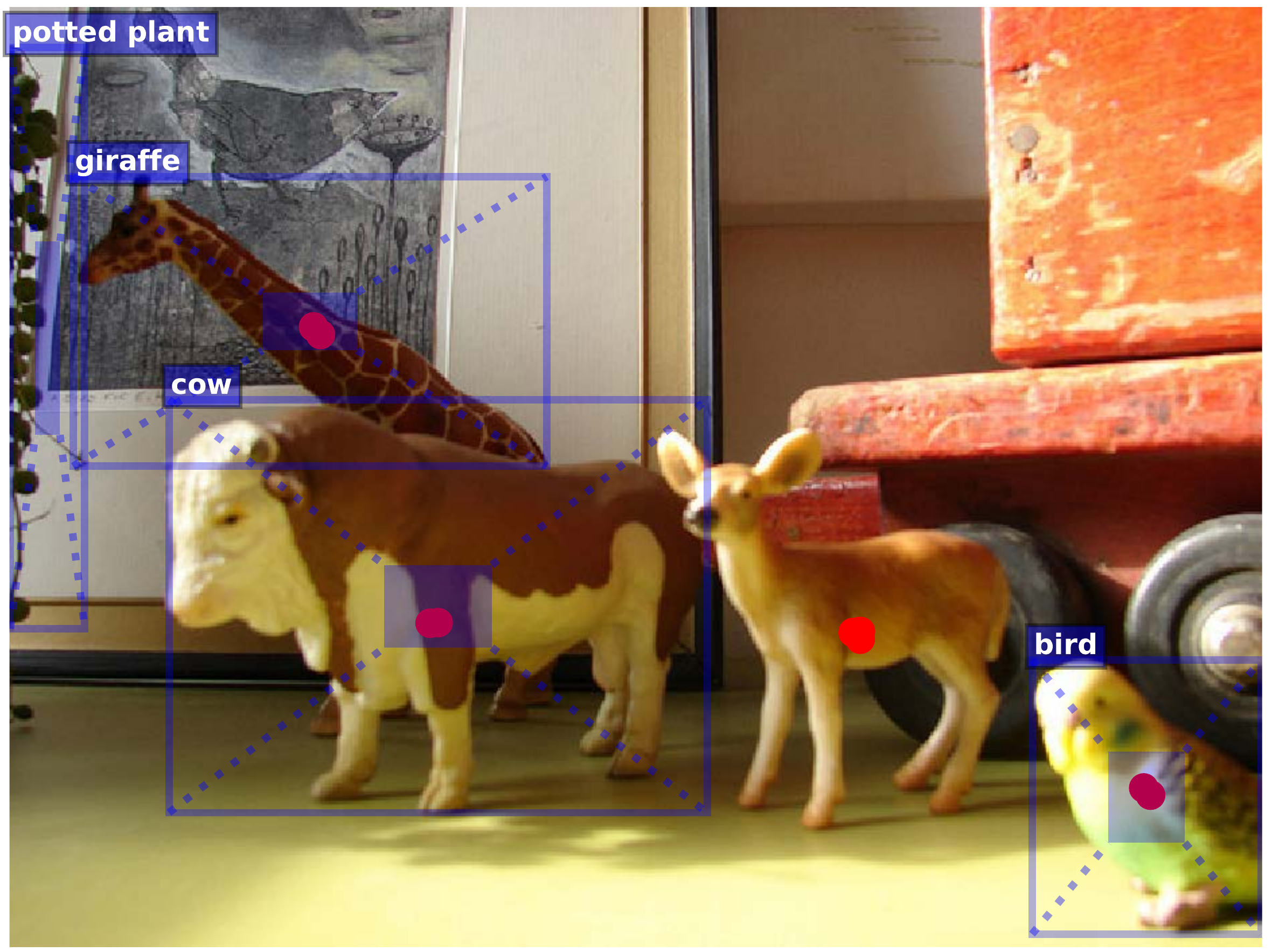}
    \label{fig7_52}
  } 
  \renewcommand\thesubfigure{(f)}
  \hspace{-0.1in}
  \subfigure[]{ 
    \includegraphics[height=0.12\textwidth,width=0.14\textheight]{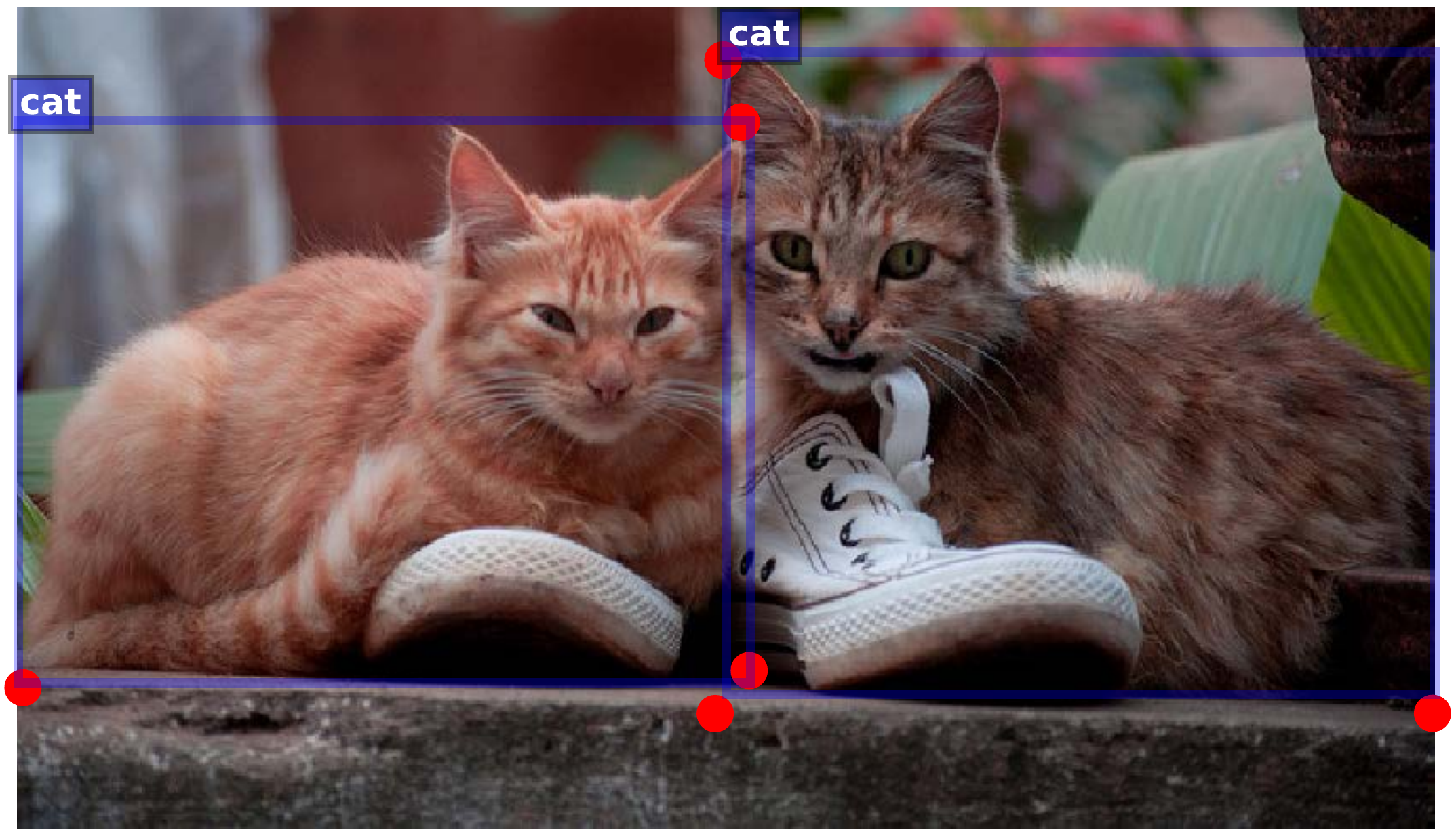}
    \label{fig7_62}
  }
  \vspace{-3ex}
  \caption{(a) and (b) show the small incorrect bounding boxes are significantly reduced by modeling center information. (c) and (d) show that the center information works for reducing medium and large incorrect bounding boxes. (e) shows the results of detecting the center keypoints without/with the center pooling. (f) shows the results of detecting the corners with corner pooling and cascade corner pooling, respectively. The blue boxes above denote the ground-truth. The red boxes and dots denote the predicted bounding boxes and keypoints, respectively.} 
  \label{qualitative}
  \vspace{-2ex}
\end{figure*}

Meanwhile, it can be seen that the most contribution comes from the small objects. For instance, CenterNet511-52 improves the AP for small objects by $5.5\%$ (single-scale) and by 
$6.4\%$ (multi-scale). As for the backbone Hourglass-104, the improvements are $6.2\%$ (single-scale) and by $8.1\%$ (multi-scale), respectively. The benefit stems from the center information modeled by the center keypoints: the smaller the scale of an incorrect bounding box is, the lower probability a center keypoint can be detected in its central region. Figure~\ref{fig7_12} and Figure~\ref{fig7_22} show some qualitative comparisons, which demonstrate the effectiveness of CenterNet in reducing small incorrect bounding boxes.

CenterNet also leads to a large improvement for reducing medium and large incorrect bounding boxes. As Table~\ref{tab1} shows, CenterNet511-104 improves the single-scale testing AP by $4.7\%$ (from $42.7\%$ to $47.4\%$) and $3.5\%$ (from $53.9\%$ to $57.4\%$), respectively. Figure~\ref{fig7_32} and Figure~\ref{fig7_42} show some qualitative comparisons for reducing medium and large incorrect bounding boxes. It is worth noting that the AR is also significantly improved, with the best performance achieved with multi-scale testing. This is because our approach removes lots of incorrect bounding boxes, which is equivalent to improving the confidence of those bounding boxes with accurate locations but lower scores.

When comparing other one-stage approaches, CenterNet511-52 reports $41.6\%$ single-scale testing AP. This achievement is already better than those using deeper models (\eg, RetinaNet800~\cite{lin2017focal} and RefineDet~\cite{zhang2018single}). The best performance of CenterNet is AP $\textbf{47.0\%}$, dramatically surpassing all the published one-stage approaches to our best knowledge.

\begin{figure*}[t]
  \centering 
  \subfigure{ 
    \includegraphics[height=0.12\textwidth,width=0.13\textheight]{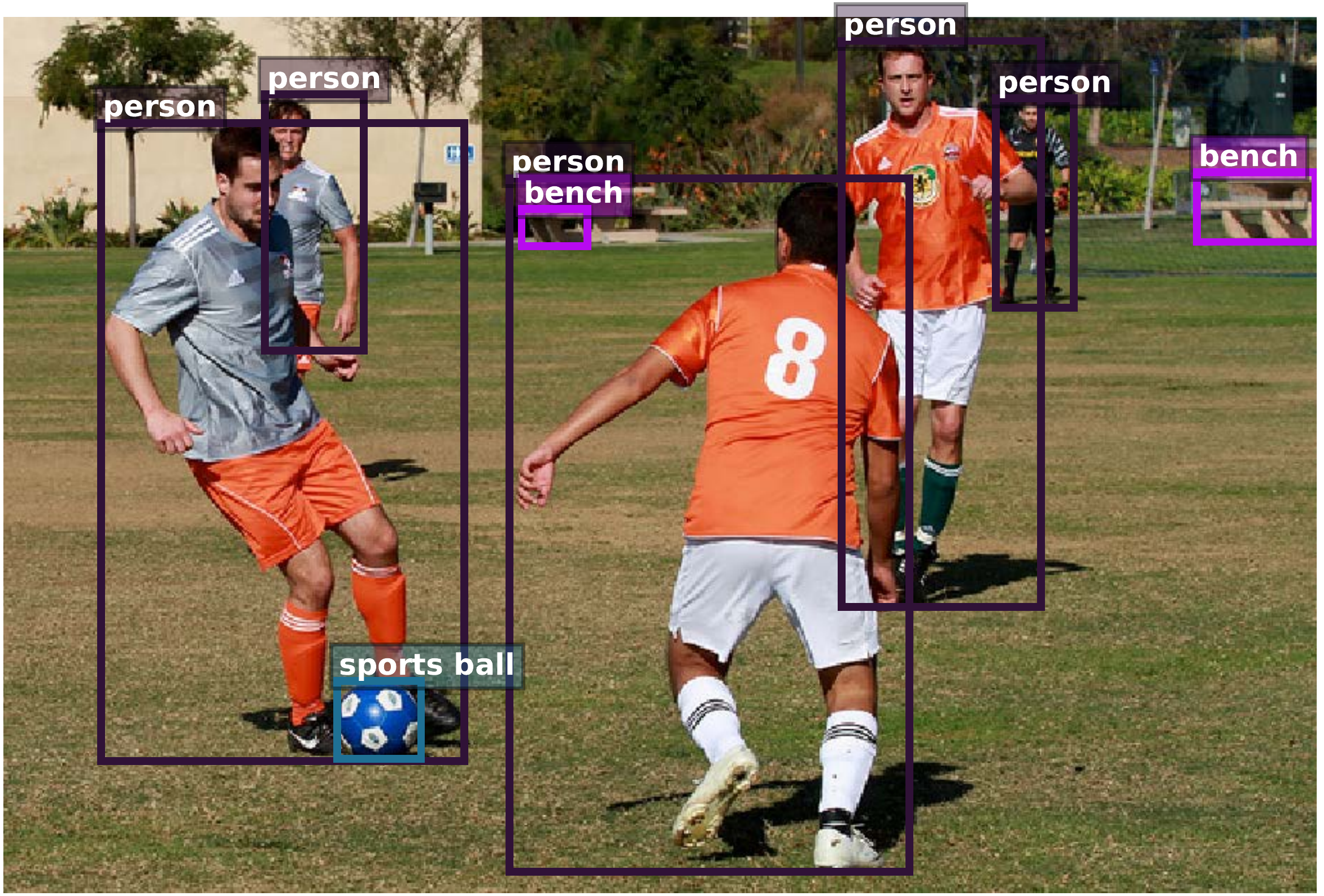}
    \label{fig8_1}
  }
  \hspace{-0.05in}
  \subfigure{ 
    \includegraphics[height=0.12\textwidth,width=0.08\textheight]{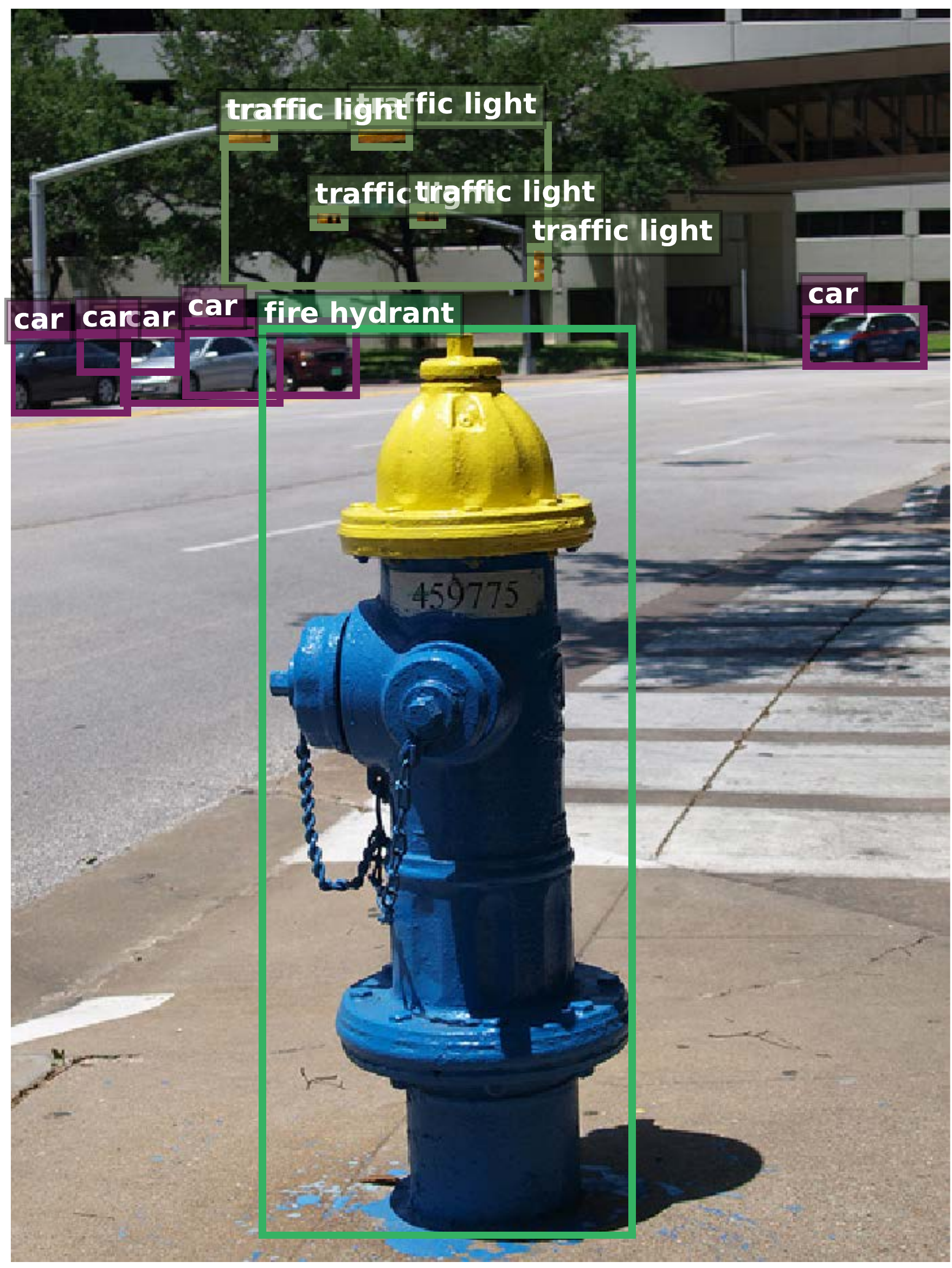}
    \label{fig8_2}
  }
  \hspace{-0.05in}
  \subfigure{ 
    \includegraphics[height=0.12\textwidth,width=0.14\textheight]{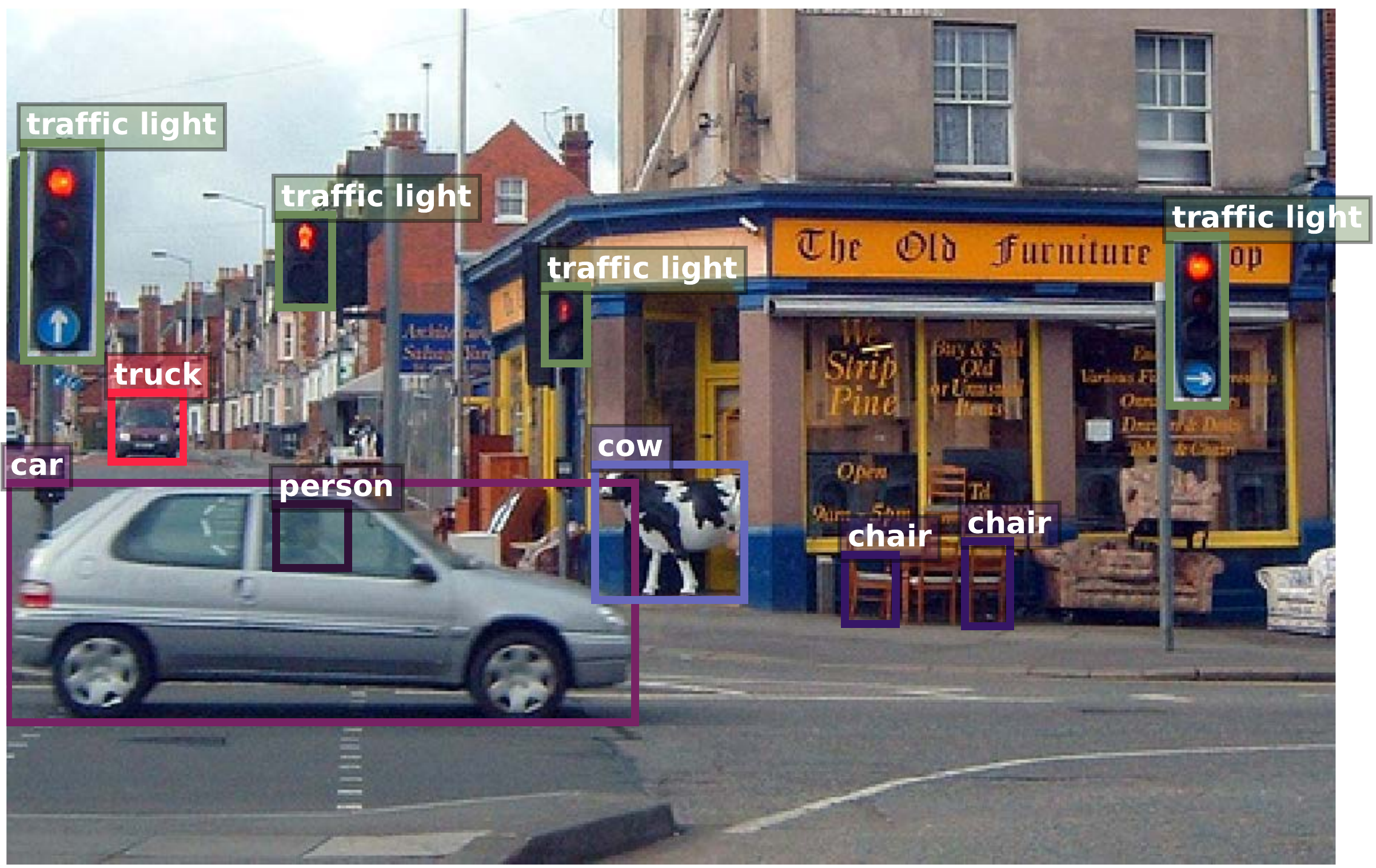}
    \label{fig8_3}
  } 
  \hspace{-0.05in}
  \subfigure{ 
    \includegraphics[height=0.12\textwidth,width=0.14\textheight]{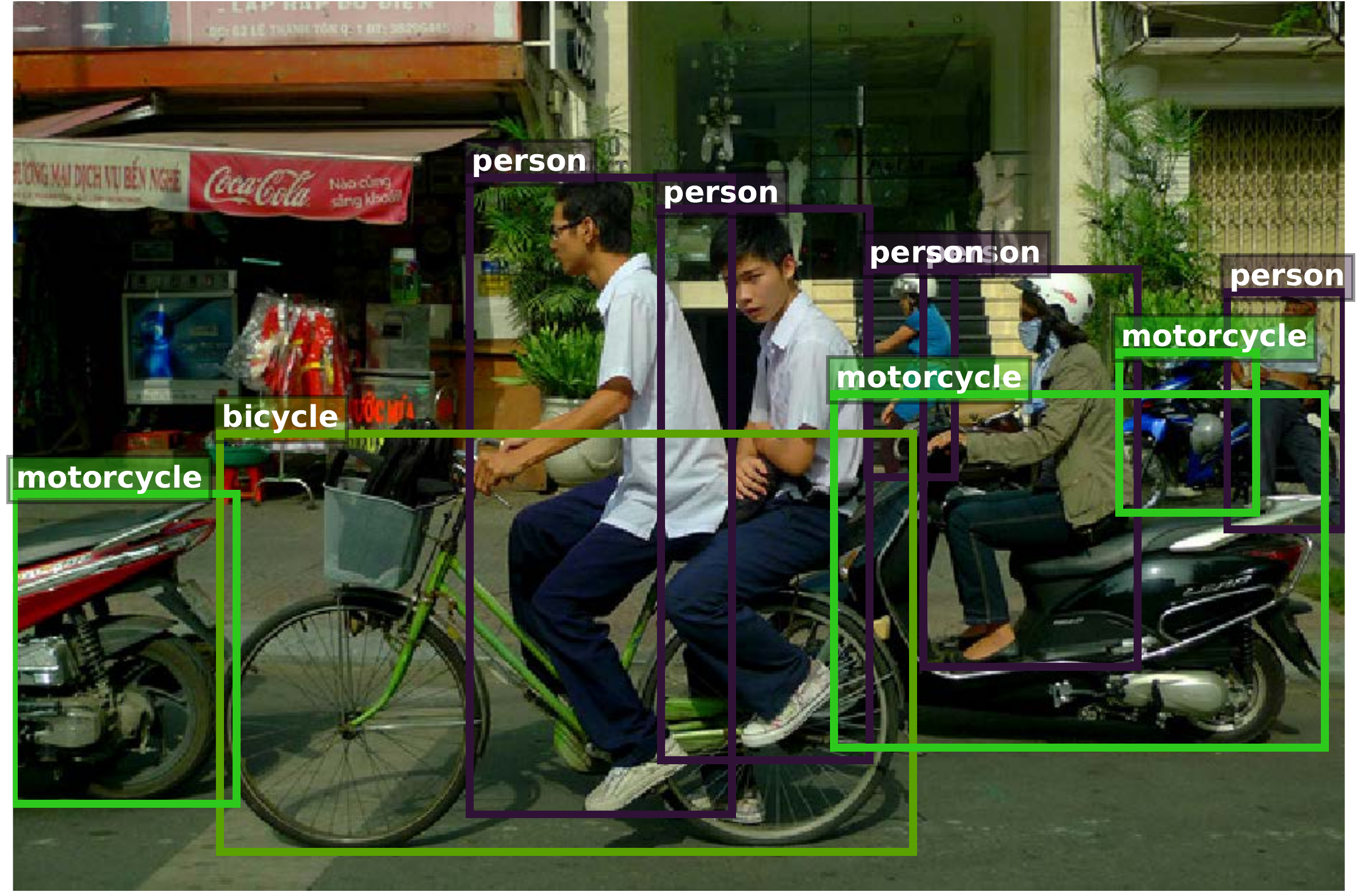}
    \label{fig8_4}
  } 
  \hspace{-0.05in}
  \subfigure{ 
    \includegraphics[height=0.12\textwidth,width=0.09\textheight]{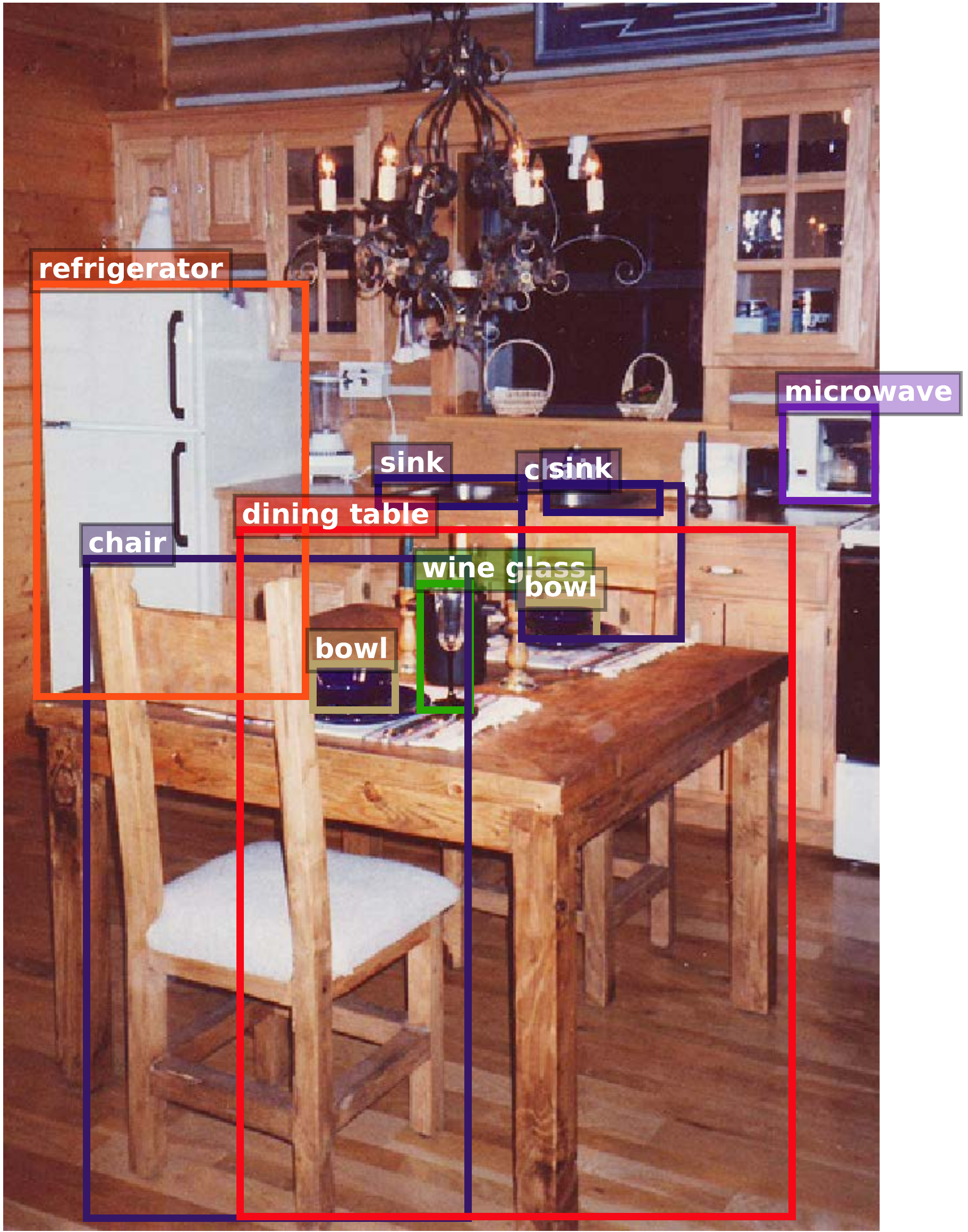}
    \label{fig8_5}
  } 
  \hspace{-0.05in}
  \subfigure{ 
    \includegraphics[height=0.125\textwidth,width=0.135\textheight]{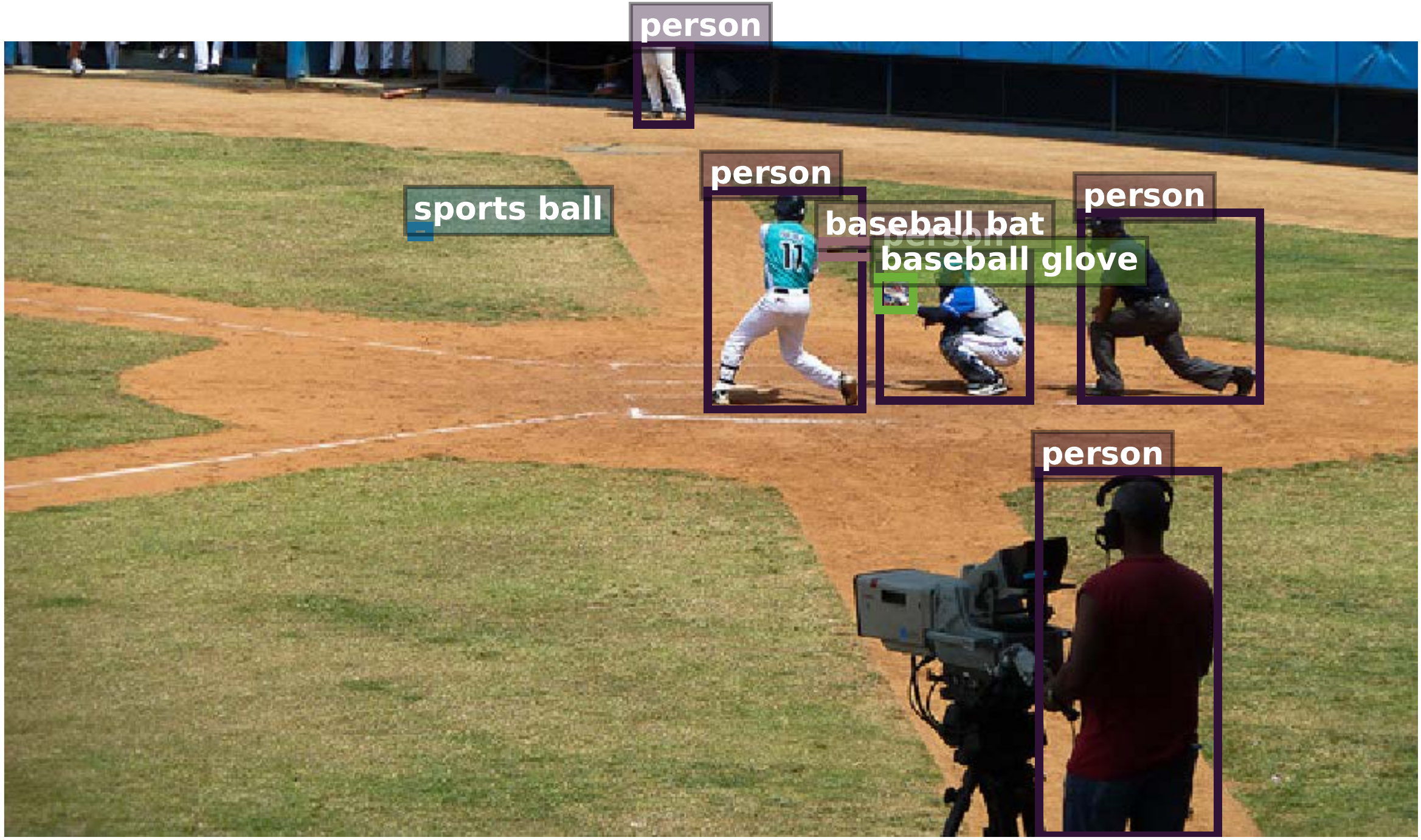}
    \label{fig8_6}
  }
  \vspace{-2ex}
  \caption{Some qualitative detection results on the MS-COCO validation dataset. Only detections with scores higher than $0.5$ are shown.} 
  \label{qualitative_detection}
  \vspace{-2ex}
\end{figure*}

At last, one can observe that the performance of CenterNet is also competitive with the two-stage approaches,~\eg,~the single-scale testing AP of CenterNet511-52 is comparable to the two-stage approach Fitness R-CNN~\cite{tychsen2018improving} ($41.6\%$~\emph{vs.}~$41.8\%$) and that of CenterNet511-104 is comparable to D-RFCN + SNIP~\cite{singh2018analysis} ($44.9\%$~\emph{vs.}~$45.7\%$), respectively. Nevertheless, it should be mentioned that two-stage approaches usually use larger resolution input images (\eg, $\sim1000\times600$), which significantly improves the detection accuracy especially for small objects. 
The multi-scale testing AP $\textbf{47.0\%}$ achieved by CenterNet511-104 closely matches the state-of-the-art AP $47.4\%$, achieved by the two-stage detector PANet~\cite{liu2018path}. We present some qualitative detection results in Figure~\ref{qualitative_detection}.

\subsection{Incorrect Bounding Box Reduction}
The AP~\cite{lin2014microsoft} metric reflects how many high quality object bounding boxes (usually $\mathrm{IoU} \geqslant 0.5$) a network can predict, but cannot directly reflect how many incorrect object bounding boxes (usually $\mathrm{IoU} \ll 0.5$) a network generates. The FD rate is a suitable metric, which reflects the proportion of the incorrect bounding boxes. Table~\ref{FDR2} shows the FD rates for CornerNet and CenterNet. CornerNet generates many incorrect bounding boxes even at $\mathrm{IoU} = 0.05$ threshold, \ie, CornerNet511-52 and CornerNet511-104 obtain $35.2\%$ and $32.7\%$ FD rate, respectively. On the other hand, CornerNet generates more small incorrect bounding boxes than medium and large incorrect bounding boxes, which reports $62.5\%$ for CornerNet511-52 and $60.3\%$ for CornerNet511-104, respectively. Our CenterNet decreases the FD rates at all criteria via exploring central regions. For instance, CenterNet511-52 and CenterNet511-104 decrease $\mathrm{FD_{5}}$ by both $4.5\%$. In addition, the FD rates for small bounding boxes decrease the most, which are $9.5\%$ by CenterNet511-52 and $9.6\%$ by CenterNet511-104, respectively. This is also the reason why the AP improvement for small objects is more prominent.
\begin{table}[tb]
\centering
\resizebox{.48\textwidth}{!}{
\begin{tabular}{|l|ccccccc|}
\hline
 Method & FD & FD$_{5}$ & FD$_{25}$ & FD$_{50}$ & FD$_{\mathrm{S}}$ & FD$_{\mathrm{M}}$ & FD$_{\mathrm{L}}$\\
\hline
\hline
 CornerNet511-52 & 40.4 & 35.2 & 39.4 & 46.7 & 62.5 & 36.9 & 28.0 \\
 CenterNet511-52 & \textbf{35.1} & \textbf{30.7} & \textbf{34.2} & \textbf{40.8} & \textbf{53.0} & \textbf{31.3} & \textbf{24.4} \\
\hline
\hline
 CornerNet511-104 & 37.8 & 32.7 & 36.8 & 43.8 & 60.3 & 33.2 & 25.1 \\
 CenterNet511-104 & \textbf{32.4} & \textbf{28.2} & \textbf{31.6} & \textbf{37.5} & \textbf{50.7} & \textbf{27.1} & \textbf{23.0} \\
\hline
\end{tabular}}
\vspace{-2ex}
\caption{Comparison of false discovery rates ($\%$) of CornerNet and CenterNet on the MS-COCO validation dataset. The results suggest CenterNet avoids a large number of incorrect bounding boxes, especially for small incorrect bounding boxes.}
\label{FDR2}
\vspace{-4ex}
\end{table}

\subsection{Inference Speed}
The proposed CenterNet explores the visual patterns within each proposed region with minimal costs. To ensure a fair comparison, we test the inference speed of both CornerNet~\cite{law2018cornernet} and CenterNet on a NVIDIA Tesla P100 GPU. We obtain that the average inference time of CornerNet511-104 is $300\mathrm{ms}$ per image and that of CenterNet511-104 is $340\mathrm{ms}$. Meanwhile, using the Hourglass-52 backbone can speed up the inference speed. Our CenterNet511-52 takes an average of $270\mathrm{ms}$ to process per image, which is faster and more accurate than CornerNet511-104.

\subsection{Ablation Study}
Our work has contributed three components, including central region exploration, center pooling and cascade corner pooling. To analyze the contribution of each individual component, an ablation study is given here. The baseline is CornerNet511-52~\cite{law2018cornernet}. We add the three components to the baseline one by one and follow the default parameter setting detailed in Section~\ref{sec:setting}. The results are given in Table~\ref{ablation}. 
\begin{table*}[tb]
\small
\centering
\begin{tabular}{|*{3}{p{0.785cm}<{\centering}}|*{6}{p{0.71cm}<{\centering}}|*{6}{p{0.71cm}<{\centering}}|}
\hline
CRE & CTP & CCP & AP & AP$_{50}$ & AP$_{75}$ & AP$_\mathrm{S}$ & AP$_\mathrm{M}$ & AP$_\mathrm{L}$ & AR$_1$ & AR$_{10}$ & AR$_{100}$ & AR$_\mathrm{S}$ & AR$_\mathrm{M}$ & AR$_\mathrm{L}$\\
\hline
\hline
 &  &  & 37.6 & 53.3 & 40.0 & 18.5 & 39.6 & 52.2 & 33.7 & 52.2 & 56.7 & 37.2 & 60.0 & 74.0 \\
 \hline
 &  & \checkmark & 38.3 & 54.2 & 40.5 & 18.6 & 40.5 & 52.2 & 34.0 & 53.0 & 57.9 & 36.6 & 60.8 & 75.8 \\
 \hline
\checkmark &  &  & 39.9 & 57.7 & 42.3 & 23.1 & 42.3 & 52.3 & 33.8 & 54.2 & 58.5 & 38.7 & 62.4 & 74.4 \\
\hline
\checkmark & \checkmark &  & 40.8 & 58.6 & 43.6 & 23.6 & 43.6 & 53.6 & 33.9 & 54.5 & 59.0 & 39.0 & 63.2 & 74.7 \\
\hline
\checkmark & \checkmark & \checkmark & \textbf{41.3} & \textbf{59.2} & \textbf{43.9} & \textbf{23.6} & \textbf{43.8} & \textbf{55.8} & \textbf{34.5} & \textbf{55.0} & \textbf{59.2} & \textbf{39.1} & \textbf{63.5} & \textbf{75.1} \\
\hline
\end{tabular}
\vspace{-2ex}
\caption{Ablation study on the major components of CenterNet511-52 on the MS-COCO validation dataset. The CRE denotes central region exploration, the CTP denotes center pooling, and the CCP denotes cascade corner pooling.}
\label{ablation}
\vspace{-2ex}
\end{table*}

\vspace{1ex}\noindent\textbf{Central region exploration.} To understand the importance of the central region exploration (see CRE in the table), we add a center heatmap branch to the baseline and use a triplet of keypoints to detect bounding boxes. For the center keypoint detection, we only use conventional convolutions. As presented in the third row in Table~\ref{ablation}, we improve the AP by $2.3\%$ (from $37.6\%$ to $39.9\%$). However, we find that the improvement for the small objects (that is $4.6$\%) is more significant than that for other object scales. The improvement for large objects is almost negligible (from $52.2\%$ to $52.3\%$). This is not surprising because, from a probabilistic point of view, the center keypoint for a small object is easier to be located than that of a large object. 

\vspace{1ex}\noindent\textbf{Center pooling.} To demonstrate the effectiveness of proposed center pooling, we then add the center pooling module to the network (see CTP in the table). The fourth row in Table~\ref{ablation} shows that center pooling improves the AP by $0.9\%$ (from $39.9\%$ to $40.8\%$). Notably, with the help of center pooling, we improve the AP for large objects by $1.4\%$ (from $52.2\%$ to $53.6\%$), which is much higher than the improvement using conventional convolutions (\ie,~$1.4\%$~\emph{vs.}~$0.1\%$). It demonstrates that our center pooling is effective in detecting center keypoints of objects, especially for large objects. Our explanation is that center pooling can extract richer internal visual patterns, and larger objects contain more accessible internal visual patterns. Figure~\ref{fig7_52} shows the results of detecting center keypoints without/with center pooling. We can see the conventional convolution fails to locate the center keypoint for the cow, but with center pooling, it successfully locates the center keypoint.
\begin{table}[!tb]
\small
\centering
\resizebox{.48\textwidth}{!}{
\begin{tabular}{|l|cccccc|}
\hline
Method & AP & AP$_{50}$ & AP$_{75}$ & AP$_{\mathrm{S}}$ & AP$_{\mathrm{M}}$ & AP$_{\mathrm{L}}$\\
\hline
\hline
CenterNet511-52 w/o GT & 41.3 & 59.2 & 43.9 & 23.6 & 43.8 & 55.8 \\
CenterNet511-52 w/ GT & \textbf{56.5} & \textbf{78.3} & \textbf{61.4} & \textbf{39.1} & \textbf{60.3} & \textbf{70.3} \\
\hline
\hline
CenterNet511-104 w/o GT & 44.8 & 62.4 & 48.2 & 25.9 & 48.9 & 58.8\\
CenterNet511-104 w/ GT & \textbf{58.1} & \textbf{78.4} & \textbf{63.9} & \textbf{40.4} & \textbf{63.0} & \textbf{72.1} \\
\hline
\end{tabular}
}
\vspace{-2ex}
\caption{Error analysis of center keypoints via using ground-truth. we replace the predicted center keypoints with the ground-truth values, the results suggest there is still room for improvement in detecting center keypoints.}
\label{Error}
\vspace{-2ex}
\end{table}

\vspace{1ex}\noindent\textbf{Cascade corner pooling.} We replace corner pooling~\cite{law2018cornernet} with cascade corner pooling to detect corners (see CCP in the table). The second row in Table~\ref{ablation} shows the results that we test on the basis of CornerNet511-52. We find that cascade corner pooling improves the AP by $0.7\%$ (from $37.6\%$ to $38.3\%$). The last row shows the results that we test on the basis of CenterNet511-52, which improves the AP by $0.5\%$ (from $40.8\%$ to $41.3\%$). The results of the second row show there is almost no change in the AP for large objects (\ie,~$52.2\%$~\emph{vs.}~$52.2\%$), but the AR is improved by $1.8\%$ (from $74.0\%$ to $75.8\%$). This suggests that cascade corner pooling can ``see" more objects due to the rich internal visual patterns in large objects, but too rich visual patterns may interfere with its perception for the boundary information, leading to many inaccurate bounding boxes. After equipping with our CenterNet, the inaccurate bounding boxes are effectively suppressed, which improves the AP for large objects by $2.2\%$ (from $53.6\%$ to $55.8\%$). Figure~\ref{fig7_62} shows the result of detecting corners with corner pooling or cascade corner pooling. We can see that cascade corner pooling can successfully locate a pair of corners for the cat on the left while corner pooling cannot.

\subsection{Error Analysis} The exploration of visual patterns within each bounding box depends on the center keypoints. In other words, once a center keypoint is missed, the proposed CenterNet would miss the visual patterns within the bounding box. To understand the importance of center keypoints, we replace the predicted center keypoints with the ground-truth values and evaluate performance on the MS-COCO validation dataset. Table~\ref{Error} shows that using the ground-truth center keypoints improves the AP from $41.3\%$ to $56.5\%$ for CenterNet511-52 and from $44.8\%$ to $58.1\%$ for CenterNet511-104, respectively. APs for small, medium and large objects are improved by $15.5\%$, $16.5\%$, and $14.5\%$ for CenterNet511-52 and $14.5\%$, $14.1\%$, and $13.3\%$ for CenterNet511-104, respectively. This demonstrates that the detection of center keypoints is far from the bottleneck.

\section{Conclusions}
\label{Conclusions}
In this paper, we propose CenterNet, which detects objects using a triplet, including one center keypoint and two corners. Our approach addresses the problem that CornerNet lacks an additional look into the cropped regions by exploring the visual patterns within each proposed region with minimal costs. In fact, this is a common defect for all one-stage approaches. As one-stage approaches remove the RoI extraction process, they cannot pay attention to internal information within cropped regions. 

{\bf An intuitive explanation of our contribution lies in that we equip a one-stage detector with the ability of two-stage approaches, with an efficient discriminator being added.} We believe that our idea of adding an extra branch for the center keypoint can be potentially generalized to other existing one-stage approaches (\eg,~SSD~\cite{liu2016ssd}). Meanwhile, some advanced training strategies~\cite{zhuscratchdet} can be used for better performance. We leave as our future work.

{\small
\bibliographystyle{ieee}
\bibliography{egbib}

\begin{thebibliography}{10}\itemsep=-1pt

\bibitem{bell2016inside}
S.~Bell, C.~Lawrence~Zitnick, K.~Bala, and R.~Girshick.
\newblock Inside-outside net: Detecting objects in context with skip pooling
  and recurrent neural networks.
\newblock In {\em Proceedings of the IEEE conference on computer vision and
  pattern recognition}, pages 2874--2883, 2016.

\bibitem{bodla2017soft}
N.~Bodla, B.~Singh, R.~Chellappa, and L.~S. Davis.
\newblock Soft-nms--improving object detection with one line of code.
\newblock In {\em Proceedings of the IEEE international conference on computer
  vision}, pages 5561--5569, 2017.

\bibitem{cai2016unified}
Z.~Cai, Q.~Fan, R.~S. Feris, and N.~Vasconcelos.
\newblock A unified multi-scale deep convolutional neural network for fast
  object detection.
\newblock In {\em European conference on computer vision}, pages 354--370.
  Springer, 2016.

\bibitem{cai2018cascade}
Z.~Cai and N.~Vasconcelos.
\newblock Cascade r-cnn: Delving into high quality object detection.
\newblock In {\em Proceedings of the IEEE conference on computer vision and
  pattern recognition}, pages 6154--6162, 2018.

\bibitem{chen2017dual}
Y.~Chen, J.~Li, H.~Xiao, X.~Jin, S.~Yan, and J.~Feng.
\newblock Dual path networks.
\newblock In {\em Advances in neural information processing systems}, pages
  4467--4475, 2017.

\bibitem{dai2016r}
J.~Dai, Y.~Li, K.~He, and J.~Sun.
\newblock R-fcn: Object detection via region-based fully convolutional
  networks.
\newblock In {\em Advances in neural information processing systems}, pages
  379--387, 2016.

\bibitem{dai2017deformable}
J.~Dai, H.~Qi, Y.~Xiong, Y.~Li, G.~Zhang, H.~Hu, and Y.~Wei.
\newblock Deformable convolutional networks.
\newblock In {\em Proceedings of the IEEE international conference on computer
  vision}, pages 764--773, 2017.

\bibitem{fu2017dssd}
C.-Y. Fu, W.~Liu, A.~Ranga, A.~Tyagi, and A.~C. Berg.
\newblock Dssd: Deconvolutional single shot detector.
\newblock {\em arXiv preprint arXiv:1701.06659}, 2017.

\bibitem{gidaris2015object}
S.~Gidaris and N.~Komodakis.
\newblock Object detection via a multi-region and semantic segmentation-aware
  cnn model.
\newblock In {\em Proceedings of the IEEE international conference on computer
  vision}, pages 1134--1142, 2015.

\bibitem{girshick2015fast}
R.~Girshick.
\newblock Fast r-cnn.
\newblock In {\em Proceedings of the IEEE international conference on computer
  vision}, pages 1440--1448, 2015.

\bibitem{girshick2014rich}
R.~Girshick, J.~Donahue, T.~Darrell, and J.~Malik.
\newblock Rich feature hierarchies for accurate object detection and semantic
  segmentation.
\newblock In {\em Proceedings of the IEEE conference on computer vision and
  pattern recognition}, pages 580--587, 2014.

\bibitem{he2017mask}
K.~He, G.~Gkioxari, P.~Doll{\'a}r, and R.~Girshick.
\newblock Mask r-cnn.
\newblock In {\em Proceedings of the IEEE international conference on computer
  vision}, pages 2961--2969, 2017.

\bibitem{he2015spatial}
K.~He, X.~Zhang, S.~Ren, and J.~Sun.
\newblock Spatial pyramid pooling in deep convolutional networks for visual
  recognition.
\newblock {\em IEEE transactions on pattern analysis and machine intelligence},
  37(9):1904--1916, 2015.

\bibitem{he2016deep}
K.~He, X.~Zhang, S.~Ren, and J.~Sun.
\newblock Deep residual learning for image recognition.
\newblock In {\em Proceedings of the IEEE conference on computer vision and
  pattern recognition}, pages 770--778, 2016.

\bibitem{hoiem2012diagnosing}
D.~Hoiem, Y.~Chodpathumwan, and Q.~Dai.
\newblock Diagnosing error in object detectors.
\newblock In {\em European conference on computer vision}, pages 340--353.
  Springer, 2012.

\bibitem{huang2017speed}
J.~Huang, V.~Rathod, C.~Sun, M.~Zhu, A.~Korattikara, A.~Fathi, I.~Fischer,
  Z.~Wojna, Y.~Song, S.~Guadarrama, et~al.
\newblock Speed/accuracy trade-offs for modern convolutional object detectors.
\newblock In {\em Proceedings of the IEEE conference on computer vision and
  pattern recognition}, pages 7310--7311, 2017.

\bibitem{jeong2017enhancement}
J.~Jeong, H.~Park, and N.~Kwak.
\newblock Enhancement of ssd by concatenating feature maps for object
  detection.
\newblock {\em arXiv preprint arXiv:1705.09587}, 2017.

\bibitem{kingma2014adam}
D.~P. Kingma and J.~Ba.
\newblock Adam: A method for stochastic optimization.
\newblock {\em Computer science}, 2014.

\bibitem{kong2017ron}
T.~Kong, F.~Sun, A.~Yao, H.~Liu, M.~Lu, and Y.~Chen.
\newblock Ron: Reverse connection with objectness prior networks for object
  detection.
\newblock In {\em Proceedings of the IEEE conference on computer vision and
  pattern recognition}, pages 5936--5944, 2017.

\bibitem{law2018cornernet}
H.~Law and J.~Deng.
\newblock Cornernet: Detecting objects as paired keypoints.
\newblock In {\em Proceedings of the European conference on computer vision},
  pages 734--750, 2018.

\bibitem{lee2017me}
H.~Lee, S.~Eum, and H.~Kwon.
\newblock Me r-cnn: Multi-expert r-cnn for object detection.
\newblock {\em arXiv preprint arXiv:1704.01069}, 2017.

\bibitem{li2019scale}
Y.~Li, Y.~Chen, N.~Wang, and Z.~Zhang.
\newblock Scale-aware trident networks for object detection.
\newblock {\em arXiv preprint arXiv:1901.01892}, 2019.

\bibitem{lin2017feature}
T.-Y. Lin, P.~Doll{\'a}r, R.~Girshick, K.~He, B.~Hariharan, and S.~Belongie.
\newblock Feature pyramid networks for object detection.
\newblock In {\em Proceedings of the IEEE conference on computer vision and
  pattern recognition}, pages 2117--2125, 2017.

\bibitem{lin2017focal}
T.-Y. Lin, P.~Goyal, R.~Girshick, K.~He, and P.~Doll{\'a}r.
\newblock Focal loss for dense object detection.
\newblock In {\em Proceedings of the IEEE international conference on computer
  vision}, pages 2980--2988, 2017.

\bibitem{lin2014microsoft}
T.-Y. Lin, M.~Maire, S.~Belongie, J.~Hays, P.~Perona, D.~Ramanan,
  P.~Doll{\'a}r, and C.~L. Zitnick.
\newblock Microsoft coco: Common objects in context.
\newblock In {\em European conference on computer vision}, pages 740--755.
  Springer, 2014.

\bibitem{liu2018path}
S.~Liu, L.~Qi, H.~Qin, J.~Shi, and J.~Jia.
\newblock Path aggregation network for instance segmentation.
\newblock In {\em Proceedings of the IEEE conference on computer vision and
  pattern recognition}, pages 8759--8768, 2018.

\bibitem{liu2016ssd}
W.~Liu, D.~Anguelov, D.~Erhan, C.~Szegedy, S.~Reed, C.-Y. Fu, and A.~C. Berg.
\newblock Ssd: Single shot multibox detector.
\newblock In {\em European conference on computer vision}, pages 21--37.
  Springer, 2016.

\bibitem{Lu2018Grid}
X.~Lu, B.~Li, Y.~Yue, Q.~Li, and J.~Yan.
\newblock Grid r-cnn.
\newblock 2018.

\bibitem{newell2016stacked}
A.~Newell, K.~Yang, and J.~Deng.
\newblock Stacked hourglass networks for human pose estimation.
\newblock In {\em European conference on computer vision}, pages 483--499.
  Springer, 2016.

\bibitem{paszke2017automatic}
A.~Paszke, S.~Gross, S.~Chintala, G.~Chanan, E.~Yang, Z.~DeVito, Z.~Lin,
  A.~Desmaison, L.~Antiga, and A.~Lerer.
\newblock Automatic differentiation in pytorch.
\newblock 2017.

\bibitem{redmon2016you}
J.~Redmon, S.~Divvala, R.~Girshick, and A.~Farhadi.
\newblock You only look once: Unified, real-time object detection.
\newblock In {\em Proceedings of the IEEE conference on computer vision and
  pattern recognition}, pages 779--788, 2016.

\bibitem{redmon2017yolo9000}
J.~Redmon and A.~Farhadi.
\newblock Yolo9000: better, faster, stronger.
\newblock In {\em Proceedings of the IEEE conference on computer vision and
  pattern recognition}, pages 7263--7271, 2017.

\bibitem{ren2015faster}
S.~Ren, K.~He, R.~Girshick, and J.~Sun.
\newblock Faster r-cnn: Towards real-time object detection with region proposal
  networks.
\newblock In {\em Advances in neural information processing systems}, pages
  91--99, 2015.

\bibitem{shen2017dsod}
Z.~Shen, Z.~Liu, J.~Li, Y.-G. Jiang, Y.~Chen, and X.~Xue.
\newblock Dsod: Learning deeply supervised object detectors from scratch.
\newblock In {\em Proceedings of the IEEE international conference on computer
  vision}, pages 1919--1927, 2017.

\bibitem{shen2017learning}
Z.~Shen, H.~Shi, R.~Feris, L.~Cao, S.~Yan, D.~Liu, X.~Wang, X.~Xue, and T.~S.
  Huang.
\newblock Learning object detectors from scratch with gated recurrent feature
  pyramids.
\newblock {\em arXiv preprint arXiv:1712.00886}, 2017.

\bibitem{shrivastava2016contextual}
A.~Shrivastava and A.~Gupta.
\newblock Contextual priming and feedback for faster r-cnn.
\newblock In {\em European conference on computer vision}, pages 330--348,
  2016.

\bibitem{shrivastava2016beyond}
A.~Shrivastava, R.~Sukthankar, J.~Malik, and A.~Gupta.
\newblock Beyond skip connections: Top-down modulation for object detection.
\newblock {\em arXiv preprint arXiv:1612.06851}, 2016.

\bibitem{singh2018analysis}
B.~Singh and L.~S. Davis.
\newblock An analysis of scale invariance in object detection snip.
\newblock In {\em Proceedings of the IEEE conference on computer vision and
  pattern recognition}, pages 3578--3587, 2018.

\bibitem{szegedy2017inception}
C.~Szegedy, S.~Ioffe, V.~Vanhoucke, and A.~A. Alemi.
\newblock Inception-v4, inception-resnet and the impact of residual connections
  on learning.
\newblock In {\em Thirty-First AAAI conference on artificial intelligence},
  2017.

\bibitem{tychsen2017denet}
L.~Tychsen-Smith and L.~Petersson.
\newblock Denet: Scalable real-time object detection with directed sparse
  sampling.
\newblock In {\em Proceedings of the IEEE international conference on computer
  vision}, pages 428--436, 2017.

\bibitem{tychsen2018improving}
L.~Tychsen-Smith and L.~Petersson.
\newblock Improving object localization with fitness nms and bounded iou loss.
\newblock In {\em Proceedings of the IEEE conference on computer vision and
  pattern recognition}, pages 6877--6885, 2018.

\bibitem{uijlings2013selective}
J.~R. Uijlings, K.~E. Van De~Sande, T.~Gevers, and A.~W. Smeulders.
\newblock Selective search for object recognition.
\newblock {\em International journal of computer vision}, 104(2):154--171,
  2013.

\bibitem{xu2018deep}
H.~Xu, X.~Lv, X.~Wang, Z.~Ren, N.~Bodla, and R.~Chellappa.
\newblock Deep regionlets for object detection.
\newblock In {\em Proceedings of the European conference on computer vision},
  pages 798--814, 2018.

\bibitem{zeng2016gated}
X.~Zeng, W.~Ouyang, B.~Yang, J.~Yan, and X.~Wang.
\newblock Gated bi-directional cnn for object detection.
\newblock In {\em European conference on computer vision}, pages 354--369.
  Springer, 2016.

\bibitem{zhang2018single}
S.~Zhang, L.~Wen, X.~Bian, Z.~Lei, and S.~Z. Li.
\newblock Single-shot refinement neural network for object detection.
\newblock In {\em Proceedings of the IEEE conference on computer vision and
  pattern recognition}, pages 4203--4212, 2018.

\bibitem{zhuscratchdet}
R.~Zhu, S.~Zhang, X.~Wang, L.~Wen, H.~Shi, L.~Bo, and T.~Mei.
\newblock Scratchdet: Training single-shot object detectors from scratch.
\newblock {\em Proceedings of the IEEE conference on computer vision and
  pattern recognition}, 2019.

\bibitem{zhu2017couplenet}
Y.~Zhu, C.~Zhao, J.~Wang, X.~Zhao, Y.~Wu, and H.~Lu.
\newblock Couplenet: Coupling global structure with local parts for object
  detection.
\newblock In {\em Proceedings of the IEEE international conference on computer
  vision}, pages 4126--4134, 2017.

\end{thebibliography}
}
\end{document}